\documentclass[journal,cspaper,compsoc]{IEEEtran}

\usepackage{graphicx}
\usepackage{balance}
\usepackage{comment}
\usepackage{cite}
\usepackage{amssymb}
\usepackage[tight,footnotesize]{subfigure}
\usepackage[active]{srcltx}
\usepackage{amsmath}
\usepackage{mathtools}

\graphicspath{{./figs/}}
\renewcommand{\baselinestretch}{0.95}
\usepackage{eurosym}
\usepackage[plain]{algorithm}
\usepackage{algorithmic}
\usepackage{multicol}
\usepackage{dsfont}
\usepackage{amsfonts}

\usepackage{cases}
\usepackage{xcolor}
\usepackage{bm}

\newlength\mylength 

\begin{document}

\title{Jointly-optimized Trajectory Generation and Camera Control for 3D Coverage Planning}

\author{Savvas~Papaioannou,~Panayiotis~Kolios,~Theocharis~Theocharides,\\~Christos~G.~Panayiotou~ and ~Marios~M.~Polycarpou 

\IEEEcompsocitemizethanks{\IEEEcompsocthanksitem The authors are with the KIOS Research and Innovation Centre of Excellence (KIOS CoE) and the Department of Electrical and Computer Engineering, University of Cyprus, Nicosia, 1678, Cyprus.\protect\\
E-mail:\texttt{\{papaioannou.savvas, pkolios, ttheocharides, christosp, mpolycar\}@ucy.ac.cy}}
\thanks{}}

\markboth{IEEE Transactions on Mobile Computing, 2025, doi:10.1109/TMC.2025.3551362}%
{Papaioannou \MakeLowercase{\textit{et al.}}: Jointly-optimized Trajectory Generation and Camera Control for 3D Coverage Planning}

\IEEEcompsoctitleabstractindextext{
\begin{abstract}
This work proposes a jointly optimized trajectory generation and camera control approach, enabling an autonomous agent, such as an unmanned aerial vehicle (UAV) operating in 3D environments, to plan and execute coverage trajectories that maximally cover the surface area of a 3D object of interest. Specifically, the UAV's kinematic and camera control inputs are jointly optimized over a rolling planning horizon to achieve complete 3D coverage of the object. The proposed controller incorporates ray-tracing into the planning process to simulate the propagation of light rays, thereby determining the visible parts of the object through the UAV's camera. This integration enables the generation of precise look-ahead coverage trajectories. The coverage planning problem is formulated as a rolling finite-horizon optimal control problem and solved using mixed-integer programming techniques. Extensive real-world and synthetic experiments validate the performance of the proposed approach.
\end{abstract}
\begin{IEEEkeywords}
Planning and control, coverage, optimization.
\end{IEEEkeywords}}

\maketitle
\IEEEdisplaynotcompsoctitleabstractindextext
\IEEEpeerreviewmaketitle

\section*{Nomenclature}
\addcontentsline{toc}{section}{Nomenclature}
\begin{IEEEdescription}[\IEEEusemathlabelsep\IEEEsetlabelwidth{$XXXXXXXX$}]
	\item[$x_t \in \mathcal{X}$] UAV state at time-step $t$.
	\item[$u_t \in \mathcal{U}$] UAV kinematic and camera control inputs at time-step $t$.
	\item[$x_{t^\prime|t}$] Predicted UAV state for time-step $t^\prime$ computed at time-step $t$.
	\item[$u_{t^\prime|t}$] Predicted control inputs for time-step $t^\prime$ computed at time-step $t$.
	\item[$\mathcal{C}_{t^\prime|t}$] Predicted UAV camera state (i.e., FOV vertices) for time-step $t^\prime$.
	\item[$\mathcal{V}_{t^\prime|t}$] The convex hull of the camera FOV at the future time-step $t^\prime$.
	\item[$\mathcal{L}=\{\Lambda_1,..,\Lambda_n\}$] The set $\mathcal{L}$ of $n$ light-rays $\Lambda_i, i \in \{1,..,n\}$.
	\item[$\mathcal{K}=\{\kappa_1,..,\kappa_{|\mathcal{K}|}\}$] The surface of the object of interest is represented as a mesh $\mathcal{K}$ composed of triangular facets $\kappa_i, i\in \{1,..,|\mathcal{K}|\}$.
	\item[$\bar{\mathcal{A}}=\{\varpi_1,..,\varpi_{|\bar{\mathcal{A}|}}\}$] The environment $\mathcal{A} \subset \mathbb{R}^3$ is discretized into a 3D grid $\bar{\mathcal{A}}$ composed of $|\bar{\mathcal{A}}|$ non-overlapping cells $\varpi_i$.
	\item[$T$] The length of the planning horizon in time-steps.
	\item[$\mathcal{M} = \Theta \times \Phi \times Z$] The set of all admissible camera configurations i.e., a combination of camera rotations $(\Theta,\Phi)$ and camera zoom levels $(Z)$.
	\item[$b_{\hat{\varpi},{\hat{\kappa}}} \in \{0,1\}$] Decision variable indicating whether there exists a light-ray that can directly trace back to facet $\kappa$ (identified by index $\hat{\kappa}$) when the UAV is located within cell $\varpi$ (identified by index $\hat{\varpi}$).
	\item[$b^{\tilde{\mathcal{A}}}_{\hat{\varpi},t^\prime|t}\in \{0,1\}$] Decision variable indicating whether at the future time-step $t^\prime$ the UAV is located within cell $\varpi$ (identified by index $\hat{\varpi}$).
	\item[$b^{\mathcal{V}}_{\hat{\kappa},\hat{m},t^\prime|t}\in \{0,1\}$] Decision variable indicating whether at time-step $t^\prime$ the facet $\kappa$ (identified by index $\hat{\kappa}$) resides within the convex-hull of the $\hat{m}_\text{th} \in |\mathcal{M}|$ camera FOV state.
	\item[$s_{\hat{m},t^\prime|t}\in \{0,1\}$] Decision variable that indicates whether at time-step $t^\prime$ the $\hat{m}_\text{th}$ camera FOV state is active.
	\item[$\bar{b}_{\hat{\kappa},\hat{m},t^\prime|t}\in \{0,1\}$] Decision variable indicating whether at time-step $t^\prime$ the facet $\kappa$ (identified by index $\hat{\kappa}$) is visible through the $\hat{m}_\text{th}$ camera FOV state.
	\item[$\mathcal{Q}(\kappa) \in \{0,1\}$] UAV memory defined as a function $\mathcal{Q}: \mathcal{K} \rightarrow \{0,1\}$, where $\mathcal{Q}(\kappa)=1$ indicates that facet $\kappa$ has been covered at some previous time-step $\tau < t$.
\end{IEEEdescription}

\section{Introduction} \label{sec:Introduction}

\IEEEPARstart{T}{he} interest in unmanned aerial vehicles (UAVs) has skyrocketed over the last decade. The latest advancements in robotics, automation, and artificial intelligence, coupled with the widespread adoption of small consumer aerial drones, have spurred unprecedented interest in UAV-based applications and services. Today, UAVs have the potential to be utilized across a wide range of application domains, including surveillance and security \cite{PapaioannouJ1,PapaioannouJ2,Saha2022,WangTMC2022,Papaioannou2022}, automated inspection \cite{Ramon2019,Bartolini2021,PapaioannouTSMC,ZachariaMED}, and emergency response missions \cite{Papaioannou2020,Papaioannou2019,Papaioannou2021b,Papaioannou2021a,9213937,10651466}.
Automated coverage planning is one of the key functionalities that can significantly enhance the autonomy of existing unmanned aerial vehicles (UAVs), enabling them to execute fully automated missions across the diverse application scenarios mentioned earlier. In coverage path planning (CPP) \cite{Cabreira2019s,Khan2017}, also referred to as coverage trajectory planning, the goal is to determine a path that allows an autonomous mobile agent to fully cover a specific area or object of interest while optimizing a particular mission objective, such as minimizing mission elapsed time.

Despite the extensive range of CPP approaches proposed in the literature, no dominant solution exists for autonomous UAV-based coverage planning in realistic 3D environments. Current state-of-the-art UAV-based CPP techniques \cite{Cabreira2019s} are primarily focused on covering planar areas (i.e., 2D terrain coverage) rather than 3D objects. Moreover, existing approaches (e.g., \cite{Torres2016,Almadhoun2019}) typically assume UAVs equipped with fixed, downward-facing cameras, thereby excluding the joint optimization and control of the UAV's kinematics and camera settings during planning. This assumption simplifies the coverage planning problem, reducing it to a path-planning problem \cite{Fazli2010,Nedjati2016}. Additionally, most CPP techniques achieve area coverage using simple geometric patterns (e.g., back-and-forth, zig-zag, or spiral motions) \cite{Tan2021}, which often lack the flexibility to generalize to 3D settings.
In this work, we argue that a practical UAV-based coverage planning solution must: (a) operate effectively in true 3D environments (i.e., producing coverage plans for 3D objects rather than planar regions), and (b) generate coverage plans in an online fashion. Modern UAVs are typically equipped with controllable cameras featuring rotation and zoom capabilities, necessitating coverage planning techniques that optimize not only the UAV's motion plan but also the control and management of the camera for optimal coverage. 

Motivated by these challenges, we propose a jointly optimized trajectory generation and camera control approach that enables a UAV agent to generate online trajectories for covering the surface area of a 3D object of interest. Specifically, we frame the coverage planning problem as a rolling finite horizon optimal control problem (FHOCP), jointly optimizing the UAV's kinematic control inputs and its camera control inputs (e.g., camera rotation and zoom levels) to produce collision-free coverage trajectories. To facilitate the generation of look-ahead coverage trajectories over a finite planning horizon, the proposed approach incorporates ray-tracing to simulate light rays captured through the UAV's camera. This enables the anticipation of the agent's behavior over the planning horizon and the generation of trajectories that account for the parts of the object visible within a finite set of planned control inputs. The key contributions of this work are as follows:

 \begin{itemize}
    \item We propose a jointly optimized trajectory generation and camera control approach for 3D coverage planning, enabling an autonomous UAV agent to efficiently generate optimal coverage trajectories for 3D objects of interest.
    \item We demonstrate how ray-tracing can be integrated into the coverage planning process to identify the visible parts of the object through the agent's camera, thereby enabling the generation of look-ahead coverage trajectories.
    \item Finally, we illustrate how this problem can be formulated as a rolling horizon optimal control problem and subsequently solved using mixed-integer programming. The effectiveness of the proposed approach is validated through extensive synthetic and real-world experiments.
\end{itemize}

This paper is structured as follows. Section~\ref{sec:Related_Work} discusses the related work on coverage path planning with autonomous agents. Section \ref{sec:system_model} outlines our modeling assumptions, Section \ref{sec:problem} formulates the problem, and Section \ref{sec:approach} discusses the proposed 3D coverage approach. Finally, Section \ref{sec:Evaluation} evaluates the proposed approach, and Section \ref{sec:conclusion} concludes the paper.

\section{Related Work}\label{sec:Related_Work}

Initial works on coverage path planning (CPP) primarily focused on ground robots operating in 2D environments \cite{Galceran2013}. Most of these approaches typically involve: (a) decomposing the area of interest into a set of non-overlapping cells, and (b) solving an optimization problem to determine the robot's path that traverses all the cells. In \cite{Acar2002sensor,Batsaikhan2013}, the authors propose coverage planning techniques that enable the robot to cover the region of interest while simultaneously constructing the cell decomposition. In \cite{Wong2003}, an exact cellular decomposition coverage planning approach based on the detection of natural landmarks is presented, while \cite{Agmon2006} introduces a spanning-tree-based CPP approach for a ground robot equipped with a fixed sensor. Additionally, \cite{wei2018coverage} investigates the CPP problem for robots with energy constraints. More recently, a coverage planning approach for 2D obstacle-cluttered environments was proposed in \cite{kan2020online}. This work introduces a hierarchical, hex-decomposition-based coverage planning algorithm that ensures complete coverage of the area of interest.

The coverage planning problem has also been extended to multi-robot systems \cite{Almadhoun2019}. For example, in \cite{hazon2006towards}, the authors propose a multi-robot spanning-tree coverage algorithm for the coverage of a bounded planar area, using multiple robots equipped with fixed sensors. In \cite{Gao2018}, the area of interest is divided among multiple robots, which then apply ant-colony optimization to determine the coverage path for their assigned regions. Similarly, \cite{Sipahioglu2010} introduces a multi-robot coverage planning approach based on the capacitated arc-routing problem, constructing coverage paths while accounting for the robots' energy capacity limitations. The authors of \cite{Rekleitis2008} proposed a boustrophedon cellular decomposition algorithm to address the coverage problem for multiple robots under communication constraints. Meanwhile, the work in \cite{Tolstaya2021} employs graph neural networks to design a multi-robot coverage system.

More recently, the coverage path planning problem has been explored for UAV-based systems and applications \cite{10557006,PapaioannouIFAC,10156482}. Specifically, the coverage of polygonal 2D areas using UAVs is investigated in \cite{Yu-Song2010,Araujo2013} and, more recently, in \cite{Xie2020}. In particular, \cite{Xie2020} proposes a UAV-based CPP approach utilizing the traveling salesman problem to cover multiple 2D polygonal regions. In \cite{TAES}, the CPP problem is formulated as an optimal control problem for a UAV operating in a bounded 2D environment, while \cite{ICUAS2022} presents a coverage planning approach tailored to cuboid-like structures. The work in \cite{avellar2015multi} employs mathematical programming techniques to command a team of UAVs for area coverage in the minimum amount of time. In \cite{apostolidis2022cooperative}, a UAV-based cooperative CPP approach is introduced, using the simulated annealing algorithm to account for each UAV's sensing and operational capabilities, initial positions, and no-fly zones. A multi-UAV CPP approach for 2D terrain coverage is proposed in \cite{Collins2021}, aiming to minimize completion time by balancing the workload across UAVs. Additionally, \cite{Perez2016} introduces a cell decomposition algorithm based on regular hexagons for multi-UAV area coverage. Finally, the approaches in \cite{apuroop2021reinforcement,Papaioannou2022CDC,lakshmanan2020complete} explore how the CPP problem can be framed as a learning problem and solved using reinforcement learning.

It is important to note here that recent advancements in 3D monitoring, communications and trajectory planning using single and multiple UAVs have led to significant developments across various applications. In \cite{torres2015high}, the authors propose a UAV-based 3D monitoring framework for accurately estimating the 3D geometric features of agricultural trees, providing valuable insights for precision agriculture and crop management optimization. In \cite{zhao2021structural}, the authors present a UAV-based system for 3D monitoring and inspection of dams using photogrammetry for damage detection. A recent survey on UAV-based 3D mapping and monitoring is available in \cite{jiang2021unmanned}. The work in \cite{hu2022disguised} introduces a 3D monitoring and video surveillance approach using fixed-wing UAVs to track suspicious mobile targets, formulating the problem as a multi-objective optimization to balance monitoring performance and power efficiency. The study in \cite{kappel2020strategies} explores various strategies and methodologies for monitoring and patrolling missions with multiple UAV agents, while \cite{outay2020applications} reviews recent developments in UAV-based monitoring applications within transportation.

Regarding multi-UAV communication and trajectory planning, \cite{zhao2021multi} investigates the problem of energy-efficient trajectory planning for UAV-based content coverage. 
A UAV-aided mobile edge computing framework is proposed in \cite{wang2020multi}, where multiple UAVs with distinct trajectories fly over the target area to support user equipment on the ground. 
In \cite{wu2018joint}, the authors design a multi-UAV-enabled wireless communication system to serve a group of ground users. The joint trajectory planning and communication problem is formulated as a mixed-integer non-convex optimization problem and solved using an efficient iterative algorithm based on block coordinate descent and successive convex optimization. \color{black} Other related work \cite{R1,R2,R3} focuses on minimizing the completion time of UAV tasks through joint communication optimization and trajectory planning, particularly considering energy costs. \color{black}

In summary, compared to existing CPP approaches, this work proposes a UAV-based coverage planning technique specifically designed for 3D environments, where both the agent and the object to be covered exist in three-dimensional space. The proposed approach jointly optimizes the UAV's kinematic and camera control inputs to generate optimal coverage trajectories. Additionally, it addresses the visibility determination problem by incorporating ray-tracing into the coverage planning process. This integration simulates the physical behavior of light rays captured by the UAV's camera, enabling the generation of look-ahead coverage trajectories over a future planning horizon. The CPP problem is formulated as a rolling finite-horizon optimal control problem (FHOCP), which is then transformed into a mixed-integer program (MIP) and solved using existing optimization tools.

\section{Preliminaries} \label{sec:system_model}

\subsection{Agent Kinematics} \label{ssec:kinematic_model}

In this work we consider a mobile UAV agent with state at time-step $t$, given by $x_t \in \mathbb{R}^6$, operating within a bounded 3D surveillance area $\mathcal{A} \subset \mathbb{R}^3$ according to the discrete-time dynamical model shown below:
\begin{equation}\label{eq:kinematics}
x_{t} = 
\begin{bmatrix}
    \text{I}_{3\times3} & \Delta t ~ \text{I}_{3\times3}\\
    \text{0}_{3\times3} & (1-a) ~ \text{I}_{3\times3}
   \end{bmatrix} x_{t-1} + 
\begin{bmatrix}
    \text{0}_{3\times3} \\
     \frac{\Delta t}{m} ~ \text{I}_{3\times3}
   \end{bmatrix} f_{t-1},
\end{equation}
where the state of the agent $x_t \in \mathcal{X}$ is composed of the 3D position, and 3D velocity vectors in Cartesian coordinates, denoted hereafter as $x_t^{\mathbf{p}}  \in \mathbb{R}^{3,1}$, and $x_t^{\mathbf{v}} \in \mathbb{R}^{3,1}$ respectively. The input vector $f_t \in \mathcal{F} \subset \mathbb{R}^{3,1}$ denotes the applied control force (i.e., the kinematic control input) which allows the agent to change its direction and speed according to the mission requirements. The parameter $a$ is used to model the air resistance, $m$ denotes the agent's mass, and $\Delta t$ is the sampling interval. Finally, the 3-by-3 square matrices  $\text{I}_{3\times3}$ and $\text{0}_{3\times3}$ denote the identity and zero matrices respectively. The proposed coverage controller, uses the kinematic model in Eq. \eqref{eq:kinematics} to construct the agent's coverage plan, which in turn is used as the reference trajectory to be tracked with a flight controller (e.g., an auto-pilot) \cite{Simone2020,Elkaim2015,Garcia2011,Yangautopilot} depending on the UAV's aerodynamical characteristics.

\subsection{Agent Camera Model} \label{ssec:sensing_model}
\begin{figure}
	\centering
	\includegraphics[width=\columnwidth]{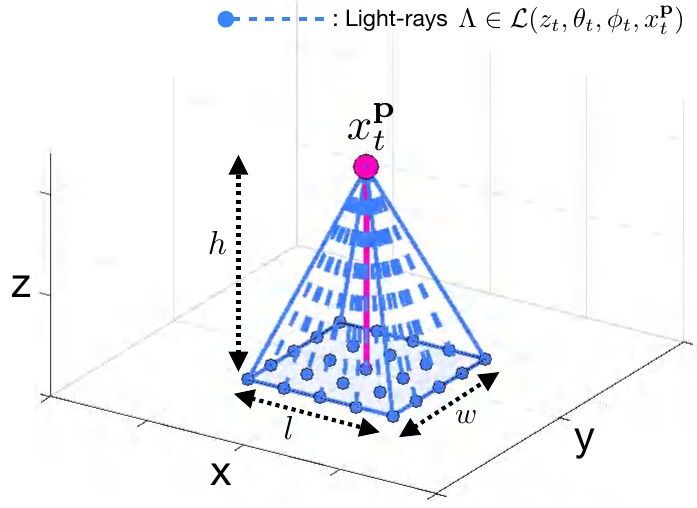}
	\caption{The camera's FOV (with horizontal and vertical FOV angles given by $\varphi_h$ and $\varphi_v$ respectively) is represented by a regular right pyramid with height $h$, and a rectangular base of dimensions $(l, w)$. At each time-step $t$ a finite set of light-rays $\mathcal{L}(z_t,\theta_t,\phi_t,x_t^{\mathbf{p}}) = \{\Lambda_1,..,\Lambda_n\}$ enter the camera's FOV (shown with the dotted line segments) as discussed in Sec. \ref{ssec:sensing_model}	}
	\label{fig:fig1}
	\vspace{-0mm}
\end{figure}

We assume that the UAV agent carries onboard a gimballed camera with rotation and zoom functionalities, which is used for viewing and sensing the surrounding environment. The shape of the camera's field-of-view (FOV) is represented in this work by a regular right pyramid which is composed of four triangular lateral faces and a rectangular base. The camera's optical center is assumed to be located at the apex which is positioned directly above the centroid of the FOV base. The FOV footprint and range is determined in this work by the parameter set $(l,w,h)$, where the pair of parameters $(l,w)$ defines the size of the FOV's rectangular base, and $h$ (i.e., the pyramid height) is the camera's observation range, as depicted in Fig. \ref{fig:fig1}. 
Subsequently, the vertices of the camera's FOV for an agent centered at the origin of the 3D cartesian coordinate frame, assuming that the camera is facing downwards, is given by the 3-by-5 matrix $\mathcal{C}_0$ as:

\begin{equation}
    \mathcal{C}_0 =
    \begin{bmatrix}
       -l/2 & l/2 & l/2  & -l/2 &0 \\
        w/2 & w/2 & -w/2 &  -w/2&0 \\
        -h  & -h  &  -h  &  -h  &0 \\
    \end{bmatrix}.
\end{equation}

As previously mentioned the agent's camera is equipped with optical zoom capabilities, which can be used to alter the FOV characteristics. Specifically, the optical zoom-in functionality increases the camera's observation range (i.e., distant objects can become observable through magnification), however this operation reduces the FOV footprint size. A particular zoom-level $z$ which belongs to the finite set of all supported zoom-levels that can be applied to the camera i.e., $z \in Z=\{z_1,...,z_{|Z|}|z_i \in \mathbb{R}, z_i \ge 1\}$  (the notation $|.|$ denotes the set cardinality), alters the FOV parameters as follows:
\begin{equation} \label{eq:zoom_function}
    h^\prime = h\times z, ~\text{and},~(l^\prime, w^\prime) = (l, w) \times z^{-1},
\end{equation}
\noindent where $(l^\prime, w^\prime, h^\prime)$ are the modified FOV parameters after applying the optical zoom level $z$. Consequently, the FOV matrix $\mathcal{C}_0$ is in fact a function of the applied zoom-level $z$ i.e., $\mathcal{C}_0(z)$. A zoom-level with value of $z=1$ does not alters the FOV characteristics, and thus describes the original FOV state without zoom. Zoom-levels with values $z > 1$, increase the observation range $h$, and scale down the FOV footprint size $(l, w)$, as shown in Eq. \eqref{eq:zoom_function}.

\begin{figure}
	\centering
	\includegraphics[width=\columnwidth]{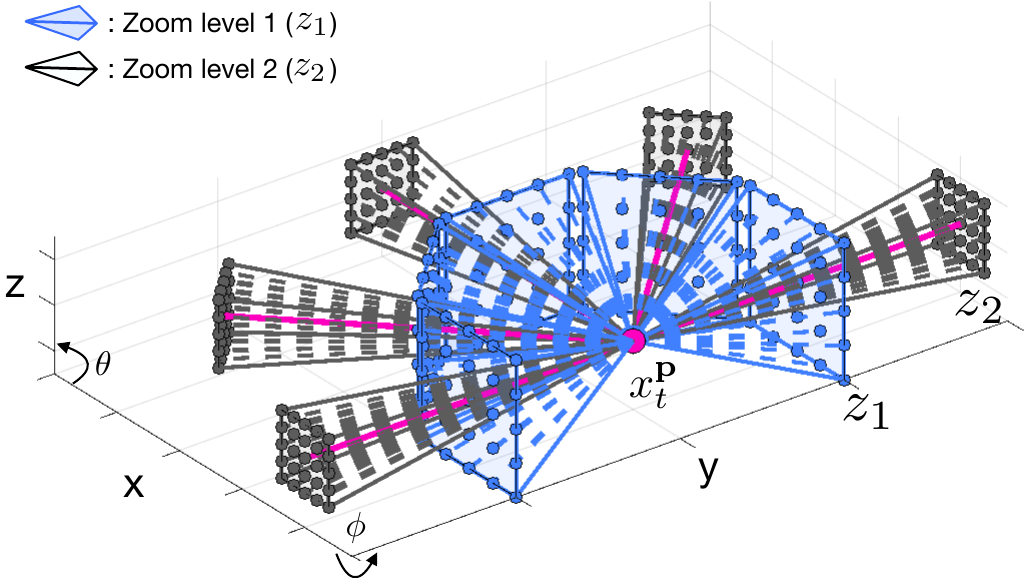}
	\caption{The figure demonstrates various configurations of the camera's FOV at time $t$ with respect to the control inputs $z_t, \theta_t, \phi_t$ (i.e., zoom-level and rotation angles). In this figure the $z_t \in \{z_1,z_2\}$, $\theta_t \in \{\frac{\pi}{2}\}$, and $\phi_t \in \{0, \frac{\pi}{4},\frac{\pi}{2},\frac{3\pi}{4},\pi\}$ as shown.}	
	\label{fig:fig2}
	\vspace{-0mm}
\end{figure}

Additionally, the camera's FOV can be rotated in 3D space by commanding the onboard camera controller to execute sequentially two elemental rotations i.e., one rotation by angle $\theta \in [0,\pi)$ around the $y-$axis, followed by a rotation $\phi \in [0,2\pi)$ around the $z-$axis. Subsequently, at time $t$ the agent can rotate and point the FOV of its camera anywhere inside the 3D surveillance space, as shown in Fig. \ref{fig:fig2}, by applying the following geometric transformation:
\begin{equation}\label{eq:fov_vertices1}
    \tilde{\mathcal{C}}_t(z_t,\theta_t,\phi_t)_i = R_{\phi}(\phi_t) R_{\theta}(\theta_t) \mathcal{C}_0(z_t)_i, \forall i \in \{1,..,5\}
\end{equation}
\noindent where $\mathcal{C}_0(z_t)_i$ is the $i_\text{th}$ column of the matrix $\mathcal{C}_0(z_t)$, $\tilde{\mathcal{C}}_t(z_t,\theta_t,\phi_t)_i$ is the rotated FOV vertex, and the rotation angles $\theta_t$ and $\phi_t$ are the camera controller's input signals at time $t$. The 3D rotation matrices $R_{\theta}(\theta_t)$ and $R_{\phi}(\phi_t)$ are further given by:
\begin{align}
 R_{\theta}(\theta_t) &=
    \begin{bmatrix}
       \text{cos}(\theta_t) & 0 & \text{sin}(\theta_t)\\
       0 & 1 & 0\\
       -\text{sin}(\theta_t) & 0 & \text{cos}(\theta_t)    
    \end{bmatrix},\\
    R_{\phi}(\phi_t) &=
    \begin{bmatrix}
       \text{cos}(\phi_t) & -\text{sin}(\phi_t) & 0\\
       \text{sin}(\phi_t) &  \text{cos}(\phi_t) & 0\\
        0 & 0 & 1
    \end{bmatrix}.
\end{align}

\noindent We should mention that the two rotation angles $\theta_t$ and $\phi_t$  take their values from the finite sets of admissible rotation angles $\Theta=\{\theta_1,..,\theta_{|\Theta|}\}$ and $\Phi=\{\phi_1,..,\phi_{|\Phi|}\}$, respectively (i.e., $\theta_t \in \Theta$ and $\phi_t \in \Phi$). To summarize, the agent's camera pose (i.e., position and orientation) in 3D space, also referred to as the camera's state, can be computed as:
\begin{equation}\label{eq:fov_vertices2}
    \mathcal{C}_t(z_t,\theta_t,\phi_t,x_t^{\mathbf{p}}) = \tilde{\mathcal{C}}_t(z_t,\theta_t,\phi_t) + x_t^{\mathbf{p}},
\end{equation}
\noindent where $x_t^{\mathbf{p}}$ is the agent's position at time-step $t$ according to Eq. \eqref{eq:kinematics}.
Finally, we consider that at each time instance $t$ a finite-set of (straight) light-rays (indicating the propagation of light) enters the camera's optical center, causing matter to be imaged (and eventually observed). The set of light-rays captured through the agent's camera with state $\mathcal{C}_t(z_t,\theta_t,\phi_t,x_t^{\mathbf{p}})$ is denoted in this work as $\mathcal{L}(z_t,\theta_t,\phi_t,x_t^{\mathbf{p}}) = \{\Lambda_1,..,\Lambda_n\}$, where $\Lambda_i, i=1,..,n$ denotes the individual light-ray in the set which is further given by the line-segment
\begin{equation}\label{eq:lightray}
    \Lambda_i = \{\lambda_i + d(x_t^{\mathbf{p}}-\lambda_i): d \in [0,1]\},
\end{equation}

\noindent 
where $x_t^{\mathbf{p}}$ is the light-ray's end point which enters the camera's optical center at time $t$ (denoted by the agent's position), $\lambda_i$ is a fixed point on the camera's FOV base denoting the ray's origin, and $d$ is a scalar. This is illustrated in Fig. \ref{fig:fig1}.

\section{Problem Formulation} \label{sec:problem}

We can define the agent's joint control signal $u_t$ at time $t$, to be composed of the agent's kinematic control input $f_t$, as well as the camera zoom-level $z_t$, and rotation angles $(\theta_t,\phi_t)$ as:
\begin{equation}
    u_t = \{f_t, z_t, \theta_t, \phi_t\},
\end{equation}
\noindent where we define $u_t^f = f_t$, $u_t^z = z_t$, $u_t^\theta = \theta_t$, and $u_t^\phi = \phi_t$. It is easy to see that, given a known agent state $\bar{x}$ at time $t=0$ i.e., $x_0=\bar{x}$, the agent's coverage trajectory (i.e., a sequence of kinematic and camera FOV states) over a finite planning horizon of length $\tilde{T}$ time-steps can be designed to meet specific criteria and goals (i.e., mission objectives and constraints) by suitably choosing the sequence of control inputs $u_t$ inside the planning horizon. In particular, the agent's coverage trajectory over the planning horizon $t \in \{1,..,\tilde{T}\}$, can be obtained as:
\begin{subequations}
\begin{align}
    &x_{t} = A^{t} x_{0} + \sum_{\tau=0}^{t-1} A^{t-\tau-1} B u_{\tau}^f, \label{eq:coverage1}\\
    &\mathcal{V}_{1:\tilde{T}} = \bigcup_{t=1}^{\tilde{T}} ~\mathcal{V}_t(u_t^z,u_t^\theta,u_t^\phi,x^{\mathbf{p}}_t) \label{eq:coverage2}
\end{align}
\end{subequations}

\noindent where Eq. \eqref{eq:coverage1} computes agent's sequence of kinematic states, with the matrices $A$, and $B$ to contain the agent's motion model parameters according to Eq. \eqref{eq:kinematics} i.e.,
\begin{equation}
A = 
\begin{bmatrix}
    \text{I}_{3\times3} & \Delta t ~ \text{I}_{3\times3}\\
    \text{0}_{3\times3} & (1-a) ~ \text{I}_{3\times3}
   \end{bmatrix}, ~
B =  
\begin{bmatrix}
    \text{0}_{3\times3} \\
     \frac{\Delta t}{m}  ~\text{I}_{3\times3}
   \end{bmatrix},
\end{equation}

\noindent and $\mathcal{V}_t$ denotes the camera's FOV convex hull generated by the FOV vertices $\mathcal{C}_t$ as computed in Eq. \eqref{eq:fov_vertices2} i.e., $\mathcal{V}_t(z_t,\theta_t,\phi_t,x_t^{\mathbf{p}}) = \triangle \left(\mathcal{C}_t(z_t,\theta_t,\phi_t,x_t^{\mathbf{p}})\right)$, where $\triangle$ denotes the convex hull operator. In essence, $\mathcal{V}_{1:\tilde{T}}$ computes the total area that the agent covered (i.e., observed) with its camera during its mission. Throughout the rest of the paper and depending on the context, both notations $\mathcal{C}_t$ and $\mathcal{V}_t$ will be used interchangably to refer to the agent's camera state (also referred to as the FOV configuration).

Suppose now that we are given an arbitrary bounded convex object of interest $\mathcal{W} \in \mathcal{A}$, with total surface area determined by its boundary $\partial\mathcal{W}$. The problem tackled in this work can be stated as follows:
 \textit{Given a sufficiently large planning horizon of length $\tilde{T}$ time-steps, find the joint control inputs $u_t, \forall t$ over the planning horizon, which optimize the coverage objective i.e., $\mathcal{G}$, and allow the mobile agent to maximally cover with its camera the surface area $\partial\mathcal{W}$ of the object of interest}.

The aforementioned coverage problem is formulated in a high-level form as the optimal control problem depicted in Problem (P1). In particular, in (P1) we are interested in the sequence of control inputs $u_0, u_1, .. , u_{\tilde{T}-1}$ which: a) optimize a given mission-specific coverage objective denoted hereafter as $\mathcal{G}$ (e.g., minimizing the mission elapsed time), and b) allow the agent to maximally cover the surface area of the object of interest in 3D. 

The constraints in Eq. \eqref{eq:P1_1}-\eqref{eq:P1_2} are due to the agent's kinematic model as described in Sec. \ref{ssec:kinematic_model}. Subsequently, the constraint in Eq. \eqref{eq:P1_3} implements the coverage functionality, which essentially states that the total surface area $\partial\mathcal{W}$ of the object of interest must reside inside the collective area covered by the agent's camera FOV during the mission, and
finally, the constraint in Eq. \eqref{eq:P1_4} makes sure that the agent avoids collisions with various obstacles $\xi \in \Xi$ in its path, including the object of interest. Lastly the constraints in Eq. \eqref{eq:P1_5} make sure that the agent's state and control inputs respect the desired operational limits. 

\begin{algorithm}
\begin{subequations}
\begin{align}
&\hspace*{-5mm}\textbf{Problem (P1):}~\texttt{Coverage Problem} &  \nonumber\\
& \hspace*{-5mm}~~~~\underset{u_0, u_1, .. , u_{\tilde{T}-1}}{\arg \min} ~\mathcal{G}(x,u)&\label{eq:objective_P1} \\
&\hspace*{-5mm}\textbf{subject to: $t \in \{1,..,\tilde{T}\}$} ~  &\nonumber\\
&\hspace*{-5mm}  x_{t} = A^{t} x_{0} + \sum_{\tau=0}^{t-1} A^{t-\tau-1} B u_{\tau}^f & \forall t\label{eq:P1_1}\\
&\hspace*{-5mm} x_0 = \bar{x} & \label{eq:P1_2}\\
&\hspace*{-5mm} \partial\mathcal{W} \in \mathcal{V}_{1:\tilde{T}} \label{eq:P1_3}\\
&\hspace*{-5mm} x_0, x^{\mathbf{p}}_t \notin \xi,~ \forall \xi \in \Xi & \forall t\label{eq:P1_4}\\
&\hspace*{-5mm} x_0, x_{t} \in \mathcal{X}, ~ u_t \in \mathcal{U}& \forall t\label{eq:P1_5}
\end{align}
\end{subequations}
\end{algorithm}

\section{Jointly-optimized Trajectory Generation and Camera Control for 3D Coverage Planning}\label{sec:approach}
The coverage planning problem discussed in the previous section and shown in problem (P1) is quite challenging to be solved efficiently. In particular, observe that a feasible solution to this problem is directly coupled with the length of the planning horizon $\tilde{T}$. If $\tilde{T}$ is too short, then no feasible solution may exist, while if $\tilde{T}$ is too long then the computational complexity increases unnecessarily.

In order to bypass this problem, (P1) is transformed into a rolling finite horizon optimal control problem (FHOCP), where at each time step: a) the current state of the agent $x_{t|t}$ is used as the initial state, and b) the  agent's control inputs  $u_{t+\tau|t}, \tau \in \{0,..,T-1\}$ are computed inside an arbitrary shorter finite horizon $T < \tilde{T}$. The first control input in the sequence $u_{t|t}$ is then applied to the agent, the agent moves to its new state, and the optimization problem is repeated for the next time step. This optimization procedure is performed iteratively until the total surface area of the object of interest is covered. The notation $x_{t^\prime|t}$ is used here to denote the future predicted agent state at time-step $t^\prime$, which is computed at time-step $t$. 

In this section we will also show how we have incorporated a ray-tracing based procedure into the proposed 3D coverage controller, in order to allow the determination of the visible parts of the scene through the agent's camera, and generate look-ahead coverage trajectories during the optimization.

\subsection{Object of Interest}\label{ssec:object_of_interest}
\begin{figure}
	\centering
	\includegraphics[width=\columnwidth]{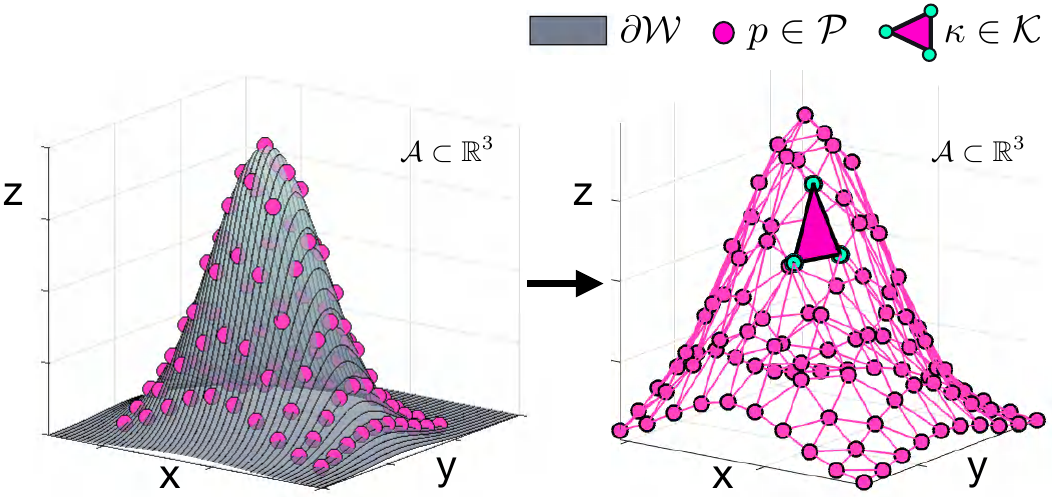}
	\caption{A 3D point-cloud representation $\mathcal{P} = \{p_1,..,p_{|\mathcal{P}|}\}$ of the object's surface $\partial \mathcal{W}$ is extracted, and then triangulated to form a triangle mesh $\mathcal{K}$ consisting of triangular facets $\kappa \in \mathcal{K}$ which need to be observed through the agent's camera.}	
	\label{fig:fig4}
	\vspace{-0mm}
\end{figure}

It is assumed that a 3D point-cloud representation of the object of interest is readily available prior to the coverage planning mission. Such representation can be acquired through a 3D scene reconstruction step \cite{Labatut2007,Dai2019} where multiple calibrated images are collected from the object of interest, from which a 3D point-cloud representation $\mathcal{P} = \{p_1,..,p_{|\mathcal{P}|}\} \in \partial \mathcal{W},~ p_i \in \mathbb{R}^3$ of the object's boundary can be extracted (i.e., points $p \in \mathcal{P}$ belong to the surface area on the object's boundary). The proposed approach takes as input the generated 3D point-cloud $\mathcal{P}$,  which is then triangulated by subdividing the object's surface into a finite set $\mathcal{K}$ of triangular facets $\kappa \in \mathbb{R}^{3\times3}$. In essence, $\mathcal{P}$ is partitioned into a finite number of non-overlapping triangular facets $\kappa \in \mathcal{K}$ as shown in Fig. \ref{fig:fig4}. Therefore, the  constraint shown in Eq. \eqref{eq:P1_3} i.e., the coverage constraint becomes: 

\begin{equation}\label{eq:discrete_coverage_con}
     \exists \mathcal{V}_{\tau} \in \mathcal{V}_{1:\tilde{T}} : \kappa \in \mathcal{V}_{\tau},~ \forall \kappa \in \mathcal{K},
\end{equation}

\noindent where with slight abuse of notation the expression $\kappa \in \mathcal{V}_\tau$ is used here to denote that the facet $\kappa$ must reside within the convex-hull of the agent's camera FOV at time-step $\tau$, and therefore we are interested in finding the agent's trajectory (i.e., kinematic and camera states) which results in the coverage of all triangle facets $\kappa \in \mathcal{K}$ which compose the object's surface area. Although, in this work we used Delaunay triangulation \cite{Lee1980,WangT2022} to create the triangle mesh $\mathcal{K}$, depending on the application scenario alternative triangulation methods can be used as well \cite{Sharp2020,Liu2009c}.

\subsection{Determining Visibility} \label{ssec:vis_con}

\begin{figure}
	\centering
	\includegraphics[width=\columnwidth]{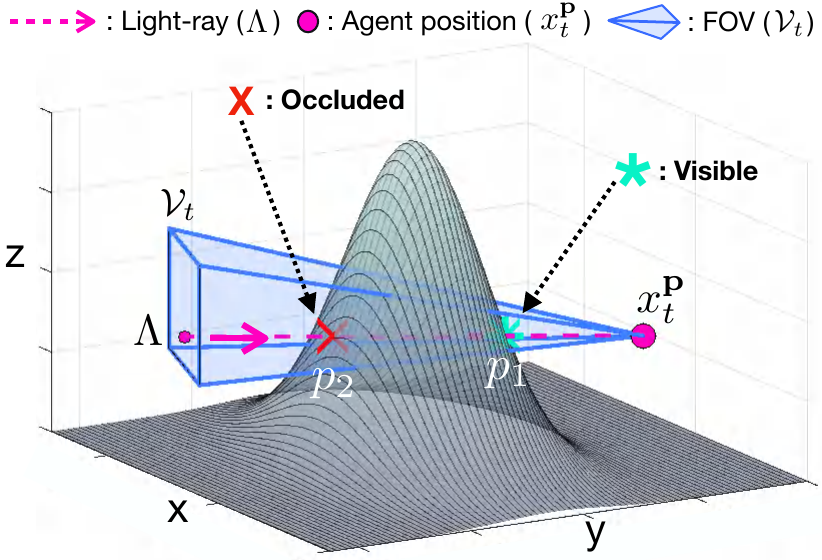}
	\caption{The figure shows that although the two points $p_1$ and $p_2$ (marked with $\ast$ and $\times$ respectively) reside inside the volume $\mathcal{V}_t$ covered by the agent's camera FOV, only point $p_1$ is visible. Specifically, in the camera configuration shown above there is no light-ray $\lambda \in \mathcal{L}(z_t,\theta_t,\phi_t,x_t^{\mathbf{p}})$ which traces back to point $p_2$ as all light-rays are blocked by the object of interest as shown.}	
	\label{fig:fig3}
	\vspace{-3mm}
\end{figure}

Although, the constraint in Eq. \eqref{eq:discrete_coverage_con} can be used to construct the agent's trajectory which covers all facets $\kappa \in \mathcal{K}$ (given a sufficiently large planning horizon $\tilde{T}$), it does not take into account the visibility problem i.e., determining whether some facet $\kappa$ which resides inside the agent's camera FOV (i.e., $\kappa \in \mathcal{V}_{\tau}$) is actually visible. An illustrative example of this problem is illustrated in Fig. \ref{fig:fig3}, which demonstrates the propagation of light-rays $\Lambda$ through the agent's camera. As shown, in this example the two points on the object's boundary $p_1$ and $p_2$, both reside inside the agent's camera FOV $\mathcal{V}_t$ at time-step $t$, however only point $p_1$ is visible given the agent location $x_t^{\mathbf{p}}$ and camera state $\mathcal{V}_t$. In fact, as illustrated in the figure point $p_2$ resides on the occluded side of the object's surface, and as a result there is no light-ray that can be traced back to point $p_2$ as all light-rays are blocked by the object's body as shown.

Intuitively, a specific part of the object's surface, represented by the triangular facet $\kappa \in \mathcal{K}$, is visible through the agent's camera (given a particular camera pose $\mathcal{C}_t(z_t,\theta_t,\phi_t,x_t^{\mathbf{p}})$) when: a) $\kappa$ resides within the camera's FOV as discussed in the previous section (i.e., $\kappa \in \mathcal{V}_t(z_t,\theta_t,\phi_t,x_t^{\mathbf{p}})$), and b) there exists a light-ray $\Lambda \in \mathcal{L}(z_t,\theta_t,\phi_t,x_t^{\mathbf{p}})$ which is not blocked and traces back to facet $\kappa$. Conversely, when no light-ray can be traced back to $\kappa$, indicates that the specific part of the object is not visible. This event might be attributed to an obstruction which blocks the propagation of light-rays and thus rendering that specific part of the object's surface unobservable. In essence, a facet is visible only when it resides inside the agent's camera FOV, and there is no occlusion intercepting the relevant light rays.  

More precisely, the notion of visibility can now be described as follows: The part of the object's surface $\kappa \in \mathcal{K}$ is visible through the agent's camera with state $\mathcal{V}_t(z_t,\theta_t,\phi_t,x_t^{\mathbf{p}})$ at time $t$ when:
\begin{equation}\label{eq:visibility_eq}
\exists \Lambda \in \mathcal{L}(z_t,\theta_t,\phi_t,x_t^{\mathbf{p}}) : \Lambda \oplus \mathcal{K} = \kappa,
\end{equation}
    
\noindent where the ray-tracing operator $\oplus$, returns the facet $\kappa \in \mathcal{K}$ which intersects last with the light-ray $\Lambda$; otherwise it returns $\emptyset$ if no facet $\kappa \in \mathcal{K}$ interests with $\Lambda$. 
In essence $\Lambda \oplus \mathcal{K}$ computes the intersections of $\Lambda$ with all the facets in the set $\mathcal{K}$, and returns (if exists) the one facet $\kappa$ which $\Lambda$ intersects just before entering the camera's optical center. The result of this operation provides the visible facet $\kappa$ (i.e., the visible part of the object's surface), since the intersecting light-ray $\Lambda$ traces back to $\kappa$, allowing the surface area captured by $\kappa$ to be imaged through the camera.

\begin{figure}
	\centering
	\includegraphics[width=\columnwidth]{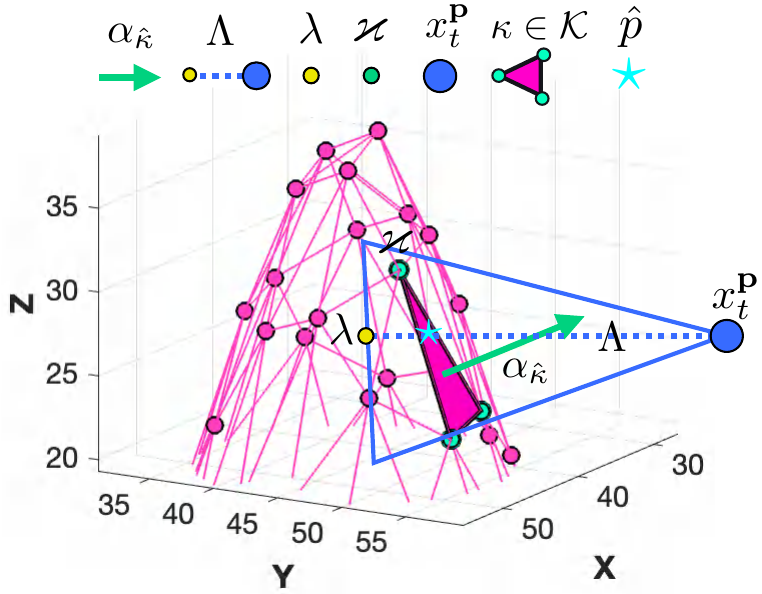}
	\caption{The figure illustrates the ray-tracing procedure discussed in Sec. \ref{ssec:vis_con} which is used in order to determine the visibility of some facet $\kappa \in \mathcal{K}$.}
	\label{fig:fig6}
	\vspace{-3mm}
\end{figure}

Let us denote the equation of the plane which contains the triangular facet $\kappa \in \mathcal{K}$ as: 
\begin{equation}\label{eq:facet_eq}
    \alpha^\top_{\hat{\kappa}} \cdot x = \beta_{\hat{\kappa}},
\end{equation}

\noindent where $\hat{\kappa} \in \{1,..,|\mathcal{K}|\}$ is the index pointing to facet $\kappa \in \mathcal{K}$, $\alpha_{\hat{\kappa}} \in \mathbb{R}^3$ is the outward normal vector to the plane containing $\kappa$, $x \in \mathbb{R}^3$, $\beta_{\hat{\kappa}} = \alpha^\top_{\hat{\kappa}} \cdot \varkappa$, $\varkappa \in \mathbb{R}^3$ is a vertex of $\kappa$, and the notation $a^\top \cdot b$ denotes the dot product of the 3D column vectors $a$ and $b$. 

That said, the operation $\Lambda \oplus \kappa$ first finds the intersection point (if exists) between the light-ray $\Lambda$ which is given by Eq. \eqref{eq:lightray} (i.e., $\Lambda = \lambda + d(x_t^{\mathbf{p}}-\lambda)$), and the plane which contains the triangular facet $\kappa$ given by Eq. \eqref{eq:facet_eq} as follows:
\begin{subequations}
\begin{align}
    &\alpha^\top_{\hat{\kappa}} \cdot [\lambda + d(x_t^{\mathbf{p}}-\lambda)] = \alpha^\top_{\hat{\kappa}} \cdot \varkappa \implies \label{eq:r1}\\
    &d = \frac{\alpha^\top_{\hat{\kappa}} \cdot (\varkappa - \lambda)}{\alpha^\top_{\hat{\kappa}} \cdot (x_t^{\mathbf{p}}-\lambda)} \label{eq:r2},
\end{align}
\end{subequations}

\noindent where Eq. \eqref{eq:r1} is the result of the substitution of $\Lambda$ for $x$ in Eq. \eqref{eq:facet_eq}, and then in Eq. \eqref{eq:r2} we solve for $d$. As a reminder in Eq. \eqref{eq:r2} $\alpha_{\hat{\kappa}}$ is the normal vector to the plane which contains facet $\kappa$, $\varkappa$ is a vertex of $\kappa$, $\lambda$ is a point on the light-ray $\Lambda$ (i.e., the origin of the ray), and finally $x_t^{\mathbf{p}}$ is the ray's end-point given by the agent's location. In essence the vector $x_t^{\mathbf{p}}-\lambda$ denotes the direction of the light-ray. An illustrative example is shown in Fig. \ref{fig:fig6}. Consequently, if the denominator of Eq. \eqref{eq:r2} is equal to zero, the light-ray $\Lambda$ and the facet $\kappa$ are parallel which results in either no visibility (i.e., $\alpha^\top_{\hat{\kappa}} \cdot (\varkappa-\lambda) \neq 0 $) or distorted view (i.e., $\alpha^\top_{\hat{\kappa}} \cdot (\varkappa-\lambda) = 0 $). Therefore, we are interested in the scenario where $\alpha^\top_{\hat{\kappa}} \cdot (x_t^{\mathbf{p}}-\lambda) \ne 0$ and thus there exists a single point intersection between the light-ray $\Lambda$ and the plane which contains the facet $\kappa$. In such scenario, the light-ray $\Lambda$ can be traced back to $\kappa$ which can render it visible through the agent's camera. However, in order for this to happen two conditions must simultaneously be satisfied: a) the value of $d$ must lie within the interval $d \in [0,1]$, otherwise the point of intersection is outside the range of the camera's FOV, and b) the point of intersection $\hat{p} = \lambda + \hat{d}(x_t^{\mathbf{p}}-\lambda)$ (where $\hat{d} \in [0,1]$ is the solution of Eq. \eqref{eq:r2}) must reside within the convex hull of the triangular facet $\kappa$ i.e., $\hat{p} \in \triangle(\kappa)$.

\begin{figure}
	\centering
	\includegraphics[scale=0.7]{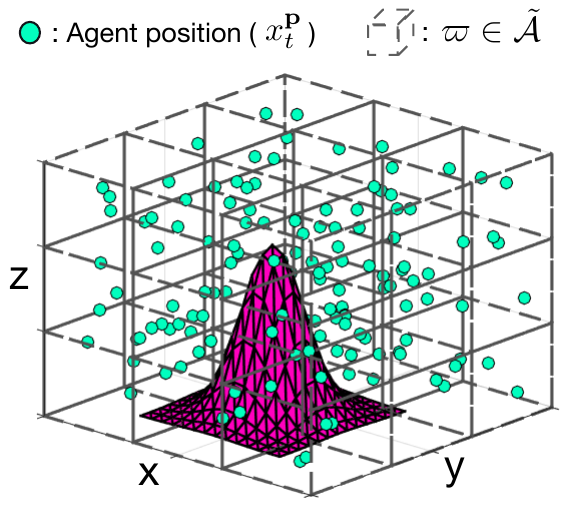}
	\caption{A set of visibility constraints $b_{\hat{\varpi},\hat{\kappa}}$ is learned through ray-tracing for each cell $\varpi \in \tilde{\mathcal{A}}$ and facet $\kappa \in \mathcal{K}$. Within each cell $\varpi$, we randomly sample the agent's position $x^{\mathbf{p}}$, along with various combinations of the control inputs $z \in Z,~ \theta \in \Theta,~ \phi \in \Phi$ to identify the visible parts of the object via the ray-tracing procedure discussed in Sec. \ref{ssec:vis_con}. }
	\label{fig:fig5}
	\vspace{-3mm}
\end{figure}

To summarize, the visibility determination process via ray-tracing discussed above must be evaluated at each time-step for all pairwise combinations of light-rays $\Lambda \in \mathcal{L}(z_t,\theta_t,\phi_t,x_t^{\mathbf{p}})$, and triangular facets $\kappa \in \mathcal{K}$, inside the planning horizon. It is easy to see that this is a computationally expensive task, which needs to be performed at each time-step. Additionally, we should mention here that, the implementation of this ray-tracing procedure introduces non-convex and non-linear constraints which are very challenging to be handled efficiently during optimization. To see this, observe that the solution to Eq. \eqref{eq:r2} depends on the unknown control inputs (i.e., $z_t, \theta_t, \phi_t$) and state of the agent (i.e., $x_t^{\mathbf{p}}$), thus rendering Eq. \eqref{eq:r2} non-convex and non-linear. 

Instead of directly implementing the ray-tracing process discussed above, in this work we employ an alternative procedure which allows the integration of ray-tracing into a mixed integer program (MIP), which can be solved using standard branch-and-bound techniques \cite{achterberg2013mixed}. To accomplish this, the ray-tracing functionality is evaluated and learned off-line on a discretized representation of the environment. Subsequently, visibility determination is approximated with a set of binary decision variables, which can easily be incorporated into a mixed integer program, thus allowing the ray-tracing functionality to be utilized during optimization.

To achieve this, the environment $\mathcal{A}$ is first discretized in space, to form a 3D grid $\tilde{\mathcal{A}}$ composed of a finite number of non-overlapping cells $\tilde{\mathcal{A}}=\{\varpi_1,..,\varpi_{|\tilde{\mathcal{A}}|}\}$, such that $\bigcup_{i=1}^{|\tilde{\mathcal{A}}|} \varpi_i= \tilde{\mathcal{A}}$. Then, within each cell $\varpi \in \tilde{\mathcal{A}}$, we randomly sample the agent's position $x^{\mathbf{p}}$, as shown in Fig. \ref{fig:fig5}, and we run the ray-tracing procedure discussed above for various joint combinations of the control inputs $z \in Z,~ \theta \in \Theta,~ \phi \in \Phi$, identifying in this way the visible parts of the object, and learning a set of state-dependent binary visibility constraints which can be embedded in a mixed integer program, and used during optimization.

Let $\tilde{\mathcal{L}}_{\hat{\varpi}}$ to denote the set of light-rays that have been obtained from the application of $N$ random combinations of the control inputs $z \in Z, \theta \in \Theta, \phi \in \Phi$, and agent states $x^{\mathbf{p}}$ which have been uniformly sampled within the cell $\varpi$ i.e.,
\begin{equation}
    \tilde{\mathcal{L}}_{\hat{\varpi}} = \bigcup_{i=1}^N \{\Lambda : \Lambda \in  \mathcal{L}^i_{\hat{\varpi}}(z_i,\theta_i,\phi_i,x_i^{\mathbf{p}}) \},
\end{equation}

\noindent where $\hat{\varpi} \in \{1,..,|\tilde{\mathcal{A}}|\}$ is the index pointing to cell $\varpi \in \tilde{\mathcal{A}}$, $\mathcal{L}^i_{\hat{\varpi}}(z_i,\theta_i,\phi_i,x_i^{\mathbf{p}})$ is the set of light-rays given by the camera pose obtained with the set of control inputs $(z_i,\theta_i,\phi_i)$, for the agent location $x_i^{\mathbf{p}}$ sampled within cell $\varpi$. We then learn the ray-tracing constraints as follows:
\begin{equation} \label{eq:vis_con1}
    b_{\hat{\varpi},{\hat{\kappa}}} = 1 \iff \exists \Lambda \in \tilde{\mathcal{L}}_{\hat{\varpi}} : \Lambda \oplus \mathcal{K} = \kappa.
\end{equation}

\noindent Thus in order to determine the visibility of facet $\kappa$, we utilize the binary variable $b_{\hat{\varpi},\hat{\kappa}}$ which is activated when there exists a light-ray $\Lambda$ which traces back to facet $\kappa$ when the agent resides within the cell $\varpi$.
In the next section we show how we have utilized the learned ray-tracing constraints in Eq. \eqref{eq:vis_con1} in the proposed 3D coverage coverage controller using mixed integer mathematical programming.

\begin{algorithm}
\begin{subequations}
\begin{align} 
&\hspace*{-2.5mm}\textbf{Problem (P2):}~\texttt{3D Coverage Controller} & \notag\\
& \hspace*{-2.5mm}~~~~\underset{u_{t|t}, u_{t+1|t}, .. , u_{t+T-1|t}}{\arg \min} ~\mathcal{G}, &  \hspace*{-10mm} \label{eq:objective_P2} \\
&\hspace*{-2.5mm}\textbf{subject to: $\tau \in \{0,..,T-1\}$} ~  &\nonumber\\
&\hspace*{-2.5mm} x_{t+\tau+1|t} = A x_{t+\tau|t} + B u_{t+\tau|t}^f, & \hspace*{-5mm} \forall \tau \label{eq:P2_1}\\
&\hspace*{-2.5mm} x_{t|t} = x_{t|t-1},&\label{eq:P2_2}\\
&\hspace*{-2.5mm} \tilde{\mathcal{C}}_{\hat{m}} =  R_\phi(u^\phi)R_\theta(u^\theta)\mathcal{C}_0(u^z),  & \hspace*{-10mm} \forall \hat{m}\label{eq:P2_3}\\
&\hspace*{-2.5mm} \mathcal{C}_{\hat{m},t+\tau+1|t} = \tilde{\mathcal{C}}_{\hat{m}}+ x_{t+\tau+1|t}^{\mathbf{p}}, & \hspace*{-5mm} \forall \hat{m}, \tau \label{eq:P2_4}\\
&\hspace*{-2.5mm} \sum_{\hat{m}=1,..,|\mathcal{M}|} s_{\hat{m},t+\tau+1|t} = 1, & \hspace*{-5mm} \forall \tau \label{eq:P2_5}\\
&\hspace*{-2.5mm} \mathcal{V}_{\hat{m},t+\tau+1|t} = \bigtriangleup(\mathcal{C}_{\hat{m},t+\tau+1|t}),  & \hspace*{-5mm} \forall \hat{m}, \tau \label{eq:P2_6}\\
&\hspace*{-2.5mm} b^{\mathcal{V}}_{\hat{\kappa},\hat{m},t+\tau+1|t} = 1 \iff  \kappa \in  \mathcal{V}_{\hat{m},t+\tau+1|t}, & \hspace*{-5mm} \forall \hat{\kappa}, \hat{m}, \tau \label{eq:P2_7}\\
&\hspace*{-2.5mm} b^{\tilde{\mathcal{A}}}_{\hat{\varpi},t+\tau+1|t} = 1 \iff  x_{t+\tau+1|t}^{\mathbf{p}} \in  \bigtriangleup(\varpi), & \hspace*{-5mm} \forall \hat{\varpi}, \tau \label{eq:P2_8}\\
&\hspace*{-2.5mm} \bar{b}_{\hat{\kappa},\hat{m},t+\tau+1|t} = s_{\hat{m},t+\tau+1|t}~\wedge & \hspace*{-10mm} \forall \hat{\kappa}, \hat{m}, \hat{\varpi}, \tau \label{eq:P2_9}\\
&\hspace*{5mm}  \Big(  b^{\mathcal{V}}_{\hat{\kappa},\hat{m},t+\tau+1|t} ~\wedge~\big[ b^{\tilde{\mathcal{A}}}_{\hat{\varpi},t+\tau+1|t}~\wedge ~ b_{\hat{\varpi},\hat{\kappa}}\big] \Big), & \hspace*{-8mm} \notag \\
&\hspace*{-2.5mm} \hat{b}_{\hat{\kappa},\hat{m},t+\tau+1|t} \leq \bar{b}_{\hat{\kappa},\hat{m},t+\tau+1|t} + \mathcal{Q}(\kappa), & \hspace*{-10mm} \forall \hat{\kappa}, \hat{m}, \tau \label{eq:P2_10}\\
&\hspace*{-2.5mm} \sum_{\tau} \sum_{\hat{m}} \hat{b}_{\hat{\kappa},\hat{m},t+\tau+1|t} \leq 1, & \hspace*{-10mm} \forall \hat{\kappa} \label{eq:P2_11}\\
&\hspace*{-2.5mm} x_{t+\tau+1|t}^{\mathbf{p}} \notin \bigtriangleup(\xi), & \hspace*{-11mm} \forall \xi \in \Xi \label{eq:P2_12}\\ 
&\hspace*{-2.5mm} x_{t+\tau+1|t} \in \mathcal{X},~ u_{t+\tau|t} \in \mathcal{U}, & \hspace*{-10mm} \notag\\ 
&\hspace*{-2.5mm}  s_{\hat{m},t+\tau+1|t},~  b^{\mathcal{V}}_{\hat{\kappa},\hat{m},t+\tau+1|t},~ b^{\tilde{\mathcal{A}}}_{\hat{\varpi},t+\tau+1|t} \in \{0,1\},   & \hspace*{-10mm}  \notag\\ 
&\hspace*{-2.5mm}  \bar{b}_{\hat{\kappa},\hat{m},t+\tau+1|t},~\hat{b}_{\hat{\kappa},\hat{m},t+\tau+1|t},~ \mathcal{Q}(\kappa) \in \{0,1\},   & \hspace*{-10mm}  \notag\\ 
&\hspace*{-2.5mm} \hat{m} \in [1,..,|\mathcal{M}|],~ \hat{\kappa} \in [1,..,|\mathcal{K}|],~\hat{\varpi} \in [1,..,|\tilde{\mathcal{A}}|]. & \hspace*{-10mm} \notag
\end{align}
\end{subequations}
\vspace{-7mm}
\end{algorithm}

\subsection{Autonomous 3D Coverage Control} \label{ssec:controller}
The proposed 3D coverage controller is shown in Problem (P2), where we have transformed the coverage problem discussed in Sec. \ref{sec:problem} into a rolling finite horizon optimal control problem, formulated as a mixed integer program (MIP). More specifically, the controller shown in Problem (P2) finds the agent's joint control inputs (i.e., kinematic and camera control inputs) $u_{t+\tau|t}, \tau \in \{0,..,T-1\}$ over the rolling horizon $T$, such that the coverage objective $\mathcal{G}$ is optimized, subject to coverage-and-visibility constraints. As we will discuss next in more detail, Problem (P2) also makes sure that the duplication of work is minimized (i.e., since the coverage mission cannot be completed within the short planning horizon $T$, the state of the mission is being tracked and saved into the agent's memory, therefore allowing the mission to progress by avoiding the duplication of work), and that  the agent avoids collisions with obstacles and the object of interest.

\subsubsection{Coverage Objective}
Let us assume that each part of the object of interest i.e., every facet $\kappa \in \mathcal{K}$, is associated with a binary variable $\hat{b}_{\hat{\kappa},\hat{m},t+\tau+1|t} \in \{0,1\}$ which is defined as below:
\begin{equation}\label{eq:cav}
 \hat{b}_{\hat{\kappa},\hat{m},t+\tau+1|t} = 
  \begin{cases} 
   1, & \text{if} ~\exists m, \tau : \kappa \in \mathcal{V}_{m,\tau} \wedge \mathcal{J}_{\mathcal{V}_{m,\tau}}(\kappa) \ne  \emptyset \\
   0, & \text{Otherwise} \\
  \end{cases}
\end{equation}

\noindent where $\tau \in \{0,..,T-1\}$, $m \in \mathcal{M}$, and $\mathcal{M}$ represents the set of all possible joint combinations of admissible camera control inputs (zoom-level and rotation angles) given by: 
\begin{equation}
    \mathcal{M} = Z \times \Theta \times \Phi,
\end{equation}
\noindent where $\times$ denotes the Cartesian product on sets, $|\mathcal{M}|$ is the total number of possible camera FOV configurations, and $\hat{m} \in \{1,..,|\mathcal{M}|\}$ is the index of the FOV configuration $m$. Thus $\mathcal{V}_{\hat{m},\tau}$ denotes the convex hull of the FOV configuration $m \in \mathcal{M}$ at time-step $t+\tau+1|t$. 
The function $\mathcal{J}_{\mathcal{V}_{\hat{m},\tau}}(\kappa)$ determines the visibility of facet $\kappa$ with respect to the FOV state $\mathcal{V}_{\hat{m},\tau}$, by implementing the ray-tracing procedure discussed in Sec. \ref{ssec:vis_con}. When facet $\kappa$ is visible at time-step $\tau$ through the $m \in \mathcal{M}$ FOV configuration $\mathcal{V}_{\hat{m},\tau}$, the function  $\mathcal{J}_{\mathcal{V}_{\hat{m},\tau}}(\kappa)$ returns the light-ray that traces back to $\kappa$, otherwise it returns the empty set when $\kappa$ is not visible (we will discuss how the functionality of $\mathcal{J}_{\mathcal{V}_{\hat{m},\tau}}(\kappa)$ was integrated into the proposed approach later in this section).

 Therefore, the binary variable $\hat{b}_{\hat{\kappa},\hat{m},t+\tau+1|t}$ determines whether facet $\kappa$ is covered and is visible through the $m$ FOV configuration at time-step $t+\tau+1|t$. We should point out here that $\mathcal{V}_{\hat{m},\tau}$ depends on the applied control inputs $u_{t+\tau|t}$, and thus the activation of $\hat{b}_{\hat{\kappa},\hat{m},t+\tau+1|t}$ can be optimized for coverage by appropriately selecting the agent's control inputs inside the planning horizon. Based on that, we can now define the coverage objective function $\mathcal{G}$ as follows:
\begin{equation}\label{eq:j1}
    \mathcal{G} = \omega\mathcal{D}(x_{t+1|t}^{\mathbf{p}},\kappa^\star)-\sum_{\hat{\kappa}=1}^{|\mathcal{K}|}\sum_{\hat{m}=1}^{|\mathcal{M}|}\sum_{\tau=0}^{T-1} \gamma(\tau) \hat{b}_{\hat{\kappa},\hat{m},t+\tau+1|t},
\end{equation}
\noindent where $\mathcal{D}(x_{t+1|t}^{\mathbf{p}},\kappa^\star) = ||x_{t+1|t}^{\mathbf{p}}-(\delta\alpha_\kappa+\kappa^\star)||_2^2$ is the squared Euclidian distance between the predicted agent position $x_{t+1|t}^{\mathbf{p}}$ and the point $\delta\alpha_{\hat{\kappa}} + \kappa^\star$, where $\kappa^\star \in \mathbb{R}^3$ is the  centroid of the nearest unobserved facet $\kappa$, $\alpha_{\hat{\kappa}}$ is the unit normal vector to the plane containing facet $\kappa$, and $\delta \in \mathbb{R}$ is a user defined positive scalar. The parameter  $\omega \in \mathbb{R}_{+}$ is a positive tuning weight, and as already discussed the binary variable $\hat{b}_{\hat{\kappa},\hat{m},t+\tau+1|t}$ is used to track the coverage events. In essence we are trying to activate $\hat{b}_{\hat{\kappa},\hat{m},t+\tau+1|t}$ inside the planning horizon for as many facets $\kappa$ as possible. This essentially maximizes the total surface area covered inside the planning horizon. The time-depended function $\gamma(\tau)$ can be used here to penalize the parts of the object that are covered later in the horizon, thus encouraging the agent to cover as many facets as quickly as possible. Observe that, the minimization of $\mathcal{G}$ is taking place inside the finite and short planning horizon $T$, which results in partial coverage of the object of interest. For this reason, $\mathcal{G}$ is optimized for subsequent time-steps $t$ for the duration of the mission, and until the object of interest is fully covered. The function $\mathcal{D}()$ is particularly useful for the mission progress (i.e., it drives the agent towards the facets that remain to be covered) in the event where although not all facets have been covered, the remaining facets cannot be reached for coverage within the planning horizon (i.e., the second term of Eq. \eqref{eq:j1} is 0). Next, we show in detail how a mixed integer program (MIP) can be formulated to optimize the 3D coverage objective function of Eq. \eqref{eq:j1}, in a rolling horizon fashion.

\subsubsection{Coverage Constraints}
Model predictive kinematic control is achieved with the constraints shown in Eq. \eqref{eq:P2_2} and Eq. \eqref{eq:P2_3}, where the kinematic control inputs $u_{t+\tau|t}^f$ are optimized inside the planning horizon according to the agent's kinematic model as discussed in Sec. \ref{ssec:kinematic_model}.

The $m \in \mathcal{M}$ camera configuration (in terms of FOV vertices) is then given in constraint Eq. \eqref{eq:P2_3} by the variable $\tilde{\mathcal{C}}_{\hat{m}}$, by applying Eq. \eqref{eq:fov_vertices1} with the $\hat{m}_\text{th}$ combination of camera control inputs (i.e., zoom-level and rotation angles respectively). Observe that the operation in Eq. \eqref{eq:P2_3} is precomputed for all $|\mathcal{M}|$ possible camera FOV states, and thus $\tilde{\mathcal{C}}_{\hat{m}}$ simply selects the camera configuration $m$.
 Subsequently, the camera FOV pose $\mathcal{C}_{\hat{m},t+\tau+1|t}$ is obtained by translating $\tilde{\mathcal{C}}_{\hat{m}}$ to the predicted agent location $x_{t+\tau+1|t}^{\mathbf{p}}$ via the constraint shown in Eq. \eqref{eq:P2_4}, where we have applied Eq. \eqref{eq:fov_vertices2} as discussed in more detail in Sec. \ref{ssec:sensing_model}.

The binary variable $s_{\hat{m},t+\tau+1|t}$ is used in Eq. \eqref{eq:P2_5} to indicate the active camera state at time-step $t+\tau+1|t$. More specifically, at each time-step only one camera state should be active. Therefore, in order to avoid the situation where more than one camera states are activated in the same time-step, we use the constraint in Eq. \eqref{eq:P2_5}. The decision variable, $s$ in Eq. \eqref{eq:P2_5} can be thought as a 2D matrix with $T$ columns and $|\mathcal{M}|$ rows, with the restriction that the sum of each column must be equal to one.

Next, the constraint in Eq. \eqref{eq:P2_6} computes the convex hull $\mathcal{V}_{\hat{m},t+\tau+1|t}$ of the FOV vertices defined by the $\hat{m}_\text{th}$ camera state $\mathcal{C}_{\hat{m},t+\tau+1|t}$ at time-step $t+\tau+1|t$. The volume enclosed within the camera FOV $\mathcal{C}_{\hat{m},t+\tau+1|t}$ is given by the system of linear inequalities:

\begin{equation}\label{eq:fov_sys}
    \alpha^\top_{\mathcal{C}_{\hat{m},t+\tau+1|t},i} \cdot x \leq \beta_{\mathcal{C}_{\hat{m},t+\tau+1|t},i}, ~ \forall i=[1,..,5],
\end{equation}

\noindent where the equation $\alpha^\top_{\mathcal{C}_{\hat{m},t+\tau+1|t},i} \cdot x = \beta_{\mathcal{C}_{\hat{m},t+\tau+1|t},i}$ describes the equation of the plane which contains face $i \in \{1,..,5\}$ of the camera's FOV. The variable $\alpha^\top_{\mathcal{C}_{\hat{m},t+\tau+1|t},i}$ is the outward normal vector to the plane which contains face $i \in \{1,..,5\}$ of the camera's FOV with state $\mathcal{C}_{\hat{m},t+\tau+1|t}$, $x \in \mathbb{R}^3$, and $\beta_{\mathcal{C}_{\hat{m},t+\tau+1|t},i}$ is a constant. Therefore, any point $x \in \mathbb{R}^3$ which satisfies all 5 inequalities shown in Eq. \eqref{eq:fov_sys} resides within the camera's FOV i.e., belongs to the convex hull defined by the camera vertices $\mathcal{C}_{\hat{m},t+\tau+1|t}$. 

The constraint in Eq. \eqref{eq:P2_7} uses the binary variable $b^{\mathcal{V}}_{\hat{\kappa},\hat{m},t+\tau+1|t}$ to decide whether at time-step $t+\tau+1|t$, the triangular facet $\kappa \in \mathcal{K}$ resides within the $\hat{m}_\text{th}$ configuration of the camera FOV defined by $\mathcal{V}_{\hat{m},t+\tau+1|t}$. This is implemented as follows:
\begin{subequations}
\begin{align} 
&  \alpha^\top_{i,\hat{m},\tau} \cdot \kappa^c + \hat{b}^{\mathcal{V}}_{i,\hat{\kappa},\hat{m},\tau}(M-\beta_{i,\hat{m},\tau}) \le M,~ \forall i,\hat{\kappa}, \hat{m}, \tau, \label{eq:v1}\\
& 5b^{\mathcal{V}}_{\hat{\kappa},\hat{m},\tau} - \sum_{i=1}^5 \hat{b}^{\mathcal{V}}_{i,\hat{\kappa},\hat{m},\tau} \le 0, ~ \forall \hat{\kappa}, \hat{m}, \tau, \label{eq:v2}
\end{align}
\end{subequations}

\noindent where we have used $\tau$ as the short notation for $t+\tau+1|t$, and thus $\alpha^\top_{\mathcal{C}_{\hat{m},t+\tau+1|t},i}$ is abbreviated as $\alpha^\top_{i,\hat{m},\tau}$ (i.e., the outward normal vector to the plane which contains face $i$ of the $m \in \mathcal{M}$ camera FOV at time-step $\tau$), and $\beta_{\mathcal{C}_{\hat{m},t+\tau+1|t},i}$ has been shortened to $\beta_{i,\hat{m},\tau}$ as shown in Eq. \eqref{eq:v1}. Moreover, in Eq. \eqref{eq:v1} the triangular facet $\kappa$ has been approximated by its center of mass $\kappa^c$, and $M$ is an arbitrary large positive constant, which is used to ensure that Eq. \eqref{eq:v1} is valid for any value of the binary decision variable $\hat{b}^{\mathcal{V}}_{i,\hat{\kappa},\hat{m},\tau}$. When $\kappa^c$ resides within $\mathcal{V}_{\hat{m},t+\tau+1|t}$, the binary variable $\hat{b}^{\mathcal{V}}_{i,\hat{\kappa},\hat{m},\tau}$ is activated for all faces i.e., $\hat{b}^{\mathcal{V}}_{i,\hat{\kappa},\hat{m},\tau}=1, \forall i=\{1,..,5\}$ in order to satisfy the constraint imposed by Eq. \eqref{eq:v1}. In such scenario, the binary variable $b^{\mathcal{V}}_{\hat{\kappa},\hat{m},t+\tau+1|t}$ (abbreviated as $b^{\mathcal{V}}_{\hat{\kappa},\hat{m},\tau}$) in Eq. \eqref{eq:v2} is activated to satisfy the constraint, which indicates that the $\hat{m}_\text{th}$ camera FOV contains facet $\kappa$ at time-step $t+\tau+1|t$. When $b^{\mathcal{V}}_{\hat{\kappa},\hat{m},\tau}=0$, then Eq. \eqref{eq:v2} is also valid as shown.

Similarly, the constraint in Eq. \eqref{eq:P2_8} uses the binary variable $b^{\tilde{\mathcal{A}}}_{\hat{\varpi},t+\tau+1|t}$ to determine if the agent's location $x_{t+\tau+1|t}^{\mathbf{p}}$ at time-step $\tau$ resides within the cell $\varpi \in \tilde{\mathcal{A}}$ as shown below:

\begin{subequations}
\begin{align} 
&  \alpha^\top_{j,\hat{\varpi}} \cdot x_{\tau}^{\mathbf{p}} + \hat{b}^{\tilde{\mathcal{A}}}_{i,\hat{\varpi},\tau}(M-\beta_{j,\hat{\varpi}}) \le M,~ \forall j, \hat{\varpi}, \tau, \label{eq:v3}\\
& 6b^{\tilde{\mathcal{A}}}_{\hat{\varpi},\tau} - \sum_{j=1}^6 \hat{b}^{\tilde{\mathcal{A}}}_{i,\hat{\varpi},\tau} \le 0, ~ \forall \hat{\varpi}, \tau, \label{eq:v4}
\end{align}
\end{subequations}

\noindent where $j \in \{1,..,6\}$ indicates the faces of the rectangular-shaped cell $\varpi \in \tilde{\mathcal{A}}$, the time-step $t+\tau+1|t$ is abbreviated as $\tau$, the parameters $(\alpha^\top_{j,\hat{\varpi}}, \beta_{j,\hat{\varpi}})$ define the coefficients of the equation of the plane which contains the $j_\text{th}$ face of the $\hat{\varpi}_\text{th}$ cell of the grid i.e., $\alpha^\top_{j,\hat{\varpi}} \cdot x = \beta_{j,\hat{\varpi}}$, and again $M$ is a large positive constant. Therefore, when the agent resides within some cell $\varpi$, the binary variable $\hat{b}^{\tilde{\mathcal{A}}}_{i,\hat{\varpi},\tau}$ is activated for all $j = \{1,..,6\}$ faces, and subsequently the binary variable $b^{\tilde{\mathcal{A}}}_{\hat{\varpi},\tau}$ is activated, thus indicating the presence of the agent within cell $\varpi$ at time-step $\tau$ i.e., $b^{\tilde{\mathcal{A}}}_{\hat{\varpi},\tau}=1$.

Subsequently, the coverage-and-visibility constraint shown in Eq. \eqref{eq:cav}, is implemented in Eq. \eqref{eq:P2_9} as the logical conjunction shown below:
\begin{align}
    &\bar{b}_{\hat{\kappa},\hat{m},t+\tau+1|t} = s_{\hat{m},t+\tau+1|t}~\wedge \notag \\
    & \Big(  b^{\mathcal{V}}_{\hat{\kappa},\hat{m},t+\tau+1|t} ~\wedge~\big[ b^{\tilde{\mathcal{A}}}_{\hat{\varpi},t+\tau+1|t}~\wedge ~ b_{\hat{\varpi},\hat{\kappa}}\big] \Big), ~ \forall \hat{\kappa}, \hat{m}, \hat{\varpi}, \tau, \notag
\end{align}

\noindent where the binary variable $\bar{b}_{\hat{\kappa},\hat{m},t+\tau+1|t}$ is activated when the centroid  $\kappa^c$ of the facet $\kappa$, is covered and is visible through the $m$ FOV configuration of the agent's camera at time-step $t+\tau+1|t$. In order for this to happen, $\kappa^c$ must first reside within the $m$ FOV configuration at time-step $t+\tau+1|t$ as indicated by the binary variable $b^{\mathcal{V}}_{\hat{\kappa},\hat{m},t+\tau+1|t}$, and at the same time the agent must reside within some cell $\varpi$ as indicated by the binary variable $b^{\tilde{\mathcal{A}}}_{\hat{\varpi},t+\tau+1|t}$ from which facet $\kappa$ is visible (i.e., there exists a light-ray captured by the agent's camera with the $m$ configuration which traces back to $\kappa$) as indicated by the binary variable $b_{\hat{\varpi},\hat{\kappa}}$ which was learned off-line as discussed in Sec. \ref{ssec:vis_con}. Finally, the binary variable $s_{\hat{m},t+\tau+1|t}$ makes sure that at time-step $t+\tau+1|t$ only the $m$ out of the $|\mathcal{M}|$ FOV configurations is active. \\

The constraint shown in Eq. \eqref{eq:P2_10} is used to ensure that the mission progress (i.e., the facets which have been covered) is being saved and tracked, thus allowing the agent to avoid the duplication of work. This is essential since in most scenarios due to the short planning horizon only partial coverage trajectories would be obtained. This necessitates the implementation of some form of memory in where the mission progress would be saved. The agent's memory is represented in this work with the function $\mathcal{Q}: \mathcal{K} \rightarrow \{0, 1\}$ which is defined as follows:
\begin{equation}
 \mathcal{Q}(\kappa) = 
  \begin{cases} 
   1, & \text{if} ~\exists ~t^\prime \leq t : \kappa ~ \text{is covered.}\\
   0, & \text{o.w} \\
  \end{cases}
\end{equation}

\noindent which is activated to indicate that facet $\kappa$ has been covered at some time-step $t^\prime \leq t$, where $t$ denotes the current time-step. The constraint in Eq. \eqref{eq:P2_10} discourages the agent from covering facets which have been already covered, thus avoiding the duplication of work. This is because the maximization of the binary variable $\hat{b}_{\hat{\kappa},\hat{m},t+\tau+1|t}$ can occur either through $\bar{b}_{\hat{\kappa},\hat{m},t+\tau+1|t}$ or through $\mathcal{Q}(\kappa)$. Thus, if $\kappa$ has been observed in the past (i.e., $\mathcal{Q}(\kappa)=1$), then there is no incentive to generate a plan which covers $\kappa$ inside the current planning horizon (i.e., there is no need to activate $\bar{b}_{\hat{\kappa},\hat{m},t+\tau+1|t}$), since $\hat{b}_{\hat{\kappa},\hat{m},t+\tau+1|t}$ has been already maximized through $\mathcal{Q}(\kappa)$. The next constraint shown in Eq. \eqref{eq:P2_11} makes sure that the agent does not plan a coverage trajectory which results in the coverage of the same facet more than once during the same planning horizon.

\begin{figure*}
	\centering
	\includegraphics[width=\textwidth]{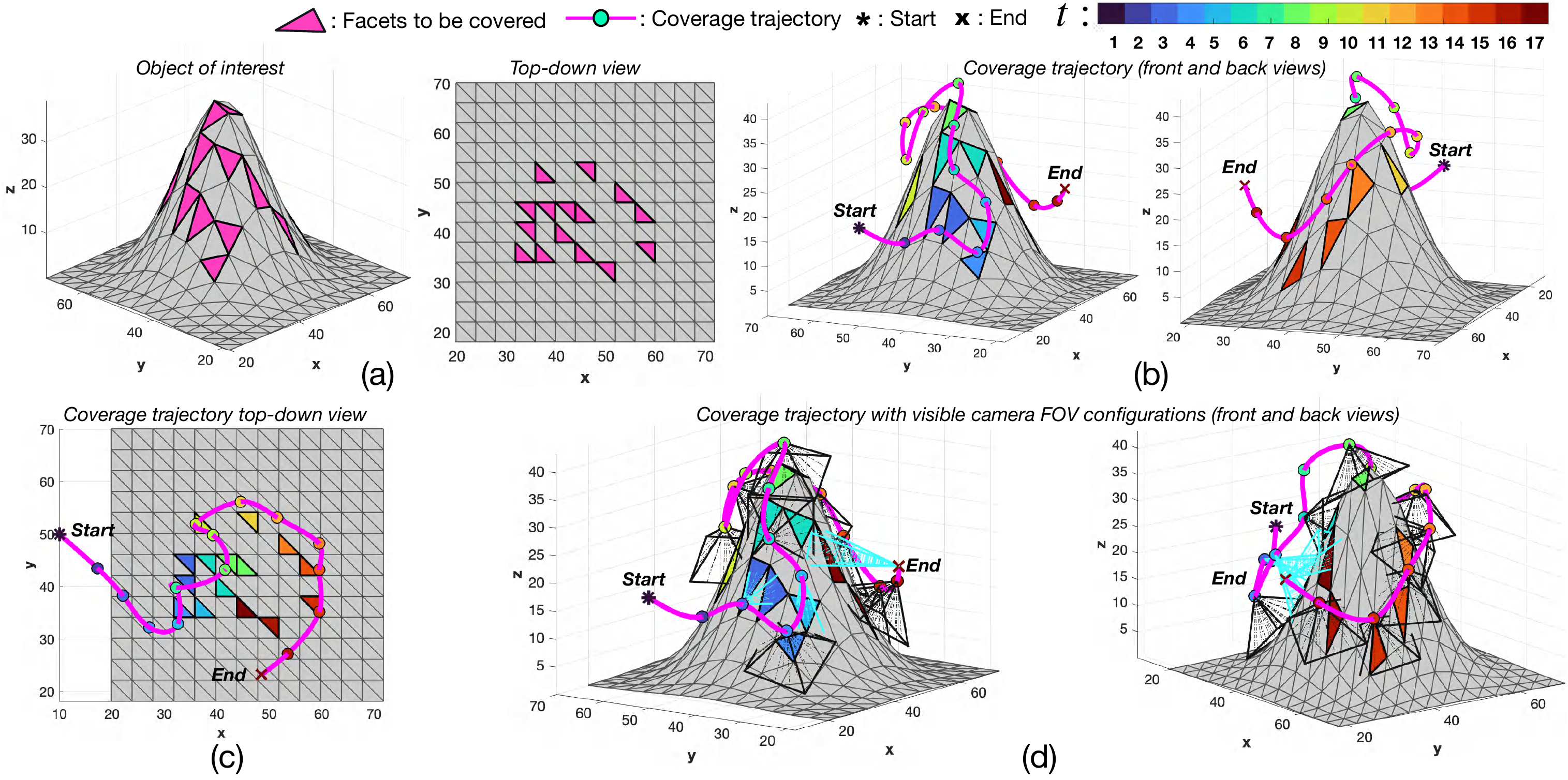}
	\caption{The figure illustrates a simulated 3D coverage planning scenario executed with the proposed approach.}
	\label{fig:res1}
	\vspace{-5mm}
\end{figure*}

Finally, the constraint shown in Eq. \eqref{eq:P2_12} makes sure that the agent avoids collisions with the obstacles in the environment, and with the object of interest. To implement this functionality we require that the agent's position $x_{t+\tau+1|t}^{\mathbf{p}}$ lies at all times outside the convex hull of all obstacles $\xi \in \Xi$ found in the environment. The procedure followed here first uses triangulation or tessellation techniques to decompose the surface area of the convex obstacle $\xi$ into a finite number of faces $f_{\hat{\xi},i}, i=\{1,..,n^{\hat{\xi}}\}$, where each face $f_{\hat{\xi},i}$ exists on the plane with equation given by:
\begin{equation} \label{eq:colision1}
    \alpha^\top_{\hat{\xi},i} \cdot x = \beta_{\hat{\xi},i}, ~ \forall i \in \{1,..,n^{\hat{\xi}}\}.
\end{equation}

\noindent where as before $\alpha_{\hat{\xi},i}$ is the normal vector to the plane which contains the $i_\text{th}$ face, $\beta_{\hat{\xi},i}$ is a constant, $x \in \mathbb{R}^3$, and $\hat{\xi} \in \{1,..,|\Xi|\}$ is the index of obstacle $\xi \in \Xi$. Subsequently, the $i_\text{th}$ plane divides the environment into two half-spaces, and the intersection of all $n^{\hat{\xi}}$ negative half-spaces forms the convex-hull of the obstacle. Therefore we use the set of constraints shown in Eq. \eqref{eq:O_1} - Eq. \eqref{eq:O_2}, and the binary variable $b^{\hat{\xi}}_{t+\tau+1|t,i}$ to make sure that a collisions with the obstacle $\xi$ is avoided at all time-steps $t+\tau+1|t$ inside the planning horizon.

\begin{algorithm}
\begin{subequations}
\begin{align}
&  \alpha^\top_{\hat{\xi},i} \cdot x_{t+\tau+1|t}^{\mathbf{p}} + M b^{\hat{\xi}}_{t+\tau+1|t,i} > \beta_{\hat{\xi},i},~\forall \hat{\xi}, \tau, i \label{eq:O_1}\\
& \sum_{i=1}^{n^{\hat{\xi}}} b^{\hat{\xi}}_{t+\tau+1|t,i} < n^{\hat{\xi}}, ~ \forall \hat{\xi}, \tau, \label{eq:O_2} 
\end{align}
\end{subequations}
\end{algorithm}

Specifically, observe from the constraint in Eq. \eqref{eq:O_1} that when the agent resides within the convex hull of obstacle $\xi$ at time-step $t+\tau+1|t$, then $ \alpha^\top_{\hat{\xi},i} \cdot x_{t+\tau+1|t}^{\mathbf{p}} \leq \beta_{\hat{\xi},i}, \forall i$, and the binary variable $b^{\hat{\xi}}_{t+\tau+1|t,i}=1, \forall i$ is activated in order to satisfy the constraint (where $M$ is a large positive constant). Thus, the agent resides outside the convex hull of the obstacle at time-step $t+\tau+1|t$, and the collision is avoided when $\exists i \in \{1,..,n^{\hat{\xi}}\} : b^{\hat{\xi}}_{t+\tau+1|t,i} = 0$. For this reason, the constraint in Eq. \eqref{eq:O_2} makes sure that the number of times $b^{\hat{\xi}}_{t+\tau+1|t,i}$ is activated at each time-step inside the planning horizon is always less than $n^{\hat{\xi}}$ as shown in Eq. \eqref{eq:O_2}.

To summarize, observe that the minimization of the coverage objective $\mathcal{G}$ shown in Eq. \eqref{eq:j1} drives the agent to select the optimal control inputs $u_{t+\tau|t}, \tau \in \{0,..,T-1\}$ inside the planning horizon which enable the activation of the binary variables $\hat{b}_{\hat{\kappa},\hat{m},t+\tau+1|t}$, which in turn indicate the coverage of certain parts of the object of interest. Full coverage is then reached by solving the problem iteratively over multiple time-steps $t$ within the duration of the mission. This is achieved by saving the mission progress inside the agent's memory, which also allows the agent to minimize the duplication of work, and complete its mission in subsequent optimization steps. Consequently, the coverage mission is terminated when all facets $\kappa \in \mathcal{K}$ are covered as indicated by the terminal condition $\sum_{\kappa \in \mathcal{K}} \mathcal{Q}(\kappa) = |\mathcal{K}|$.

\section{Evaluation} \label{sec:Evaluation}
The evaluation of the proposed 3D coverage approach is divided into two parts. In the first part we conduct a thorough analysis of the proposed approach using synthetic tests, showcasing its performance in a variety of simulated scenarios. In particular, we show the behavior of the proposed controller in a simulated 3D coverage scenario, demonstrating the role of ray-tracing during the coverage mission, and then we investigate the agent's coverage trajectory using various optimization objectives. In the second part of the evaluation, we illustrate the performance of the proposed approach in a real-world coverage mission.

\begin{figure}
	\centering
	\includegraphics[width=\columnwidth]{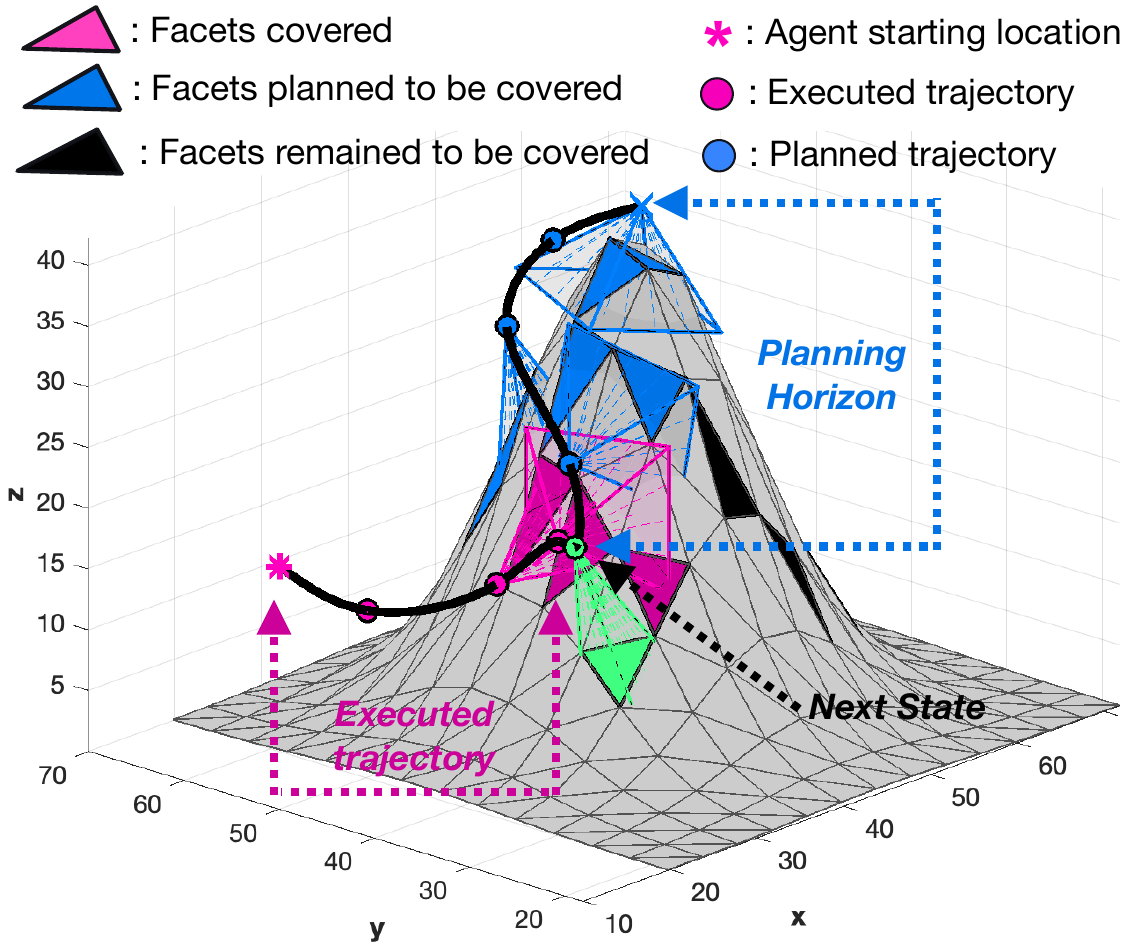}
	\caption{The figure shows the planned coverage trajectory over a horizon length of 5 time-steps, for time instances $t+\tau+1|t, t=4, \tau \in [0,..,4]$.}
	\label{fig:res2}
	\vspace{-3mm}
\end{figure}

\subsection{Synthetic Experiments}\label{ssec:simtest}
For the evaluation of the proposed approach we have used the following simulation setup based on the specifications of the DJI Mavic Enterprise drone which was utilized in our field tests. The agent's kinematic model parameters $\Delta t$, $\alpha$, and $m$ have been set to $1$s, 0.2, and $1.1$kg respectively. The agent's velocity $x_t^{\mathbf{v}}$ and motion control input $f_t$ are bounded in each dimension inside the intervals $[-15,15]$m/s and $[-10,10]$N respectively. The parameter set $(l,w,h)$ of the camera FOV has been set as $(9.5,9.5,8)$m, with $|Z|=2$ zoom-levels i.e., $Z=\{1, 2\}$, and $n=50$ light-rays. The rotation angles $\theta$ and $\phi$ take their values from the sets $\Theta=\{30, 90, 150\}$deg, and  $\Phi=\{30, 105, 180, 255, 330\}$deg respectively. This enables the camera FOV to take one out of $|M|=30$ possible configurations (i.e., 15 configurations per zoom-level). The volume of the 3D environment $\mathcal{A}$ is equal to $100^3$m, which was uniformly discretized into $|\tilde{\mathcal{A}}|=1000$ non-overlapping cells in order to determine the visibility constraints with the procedure described in Sec. \ref{ssec:vis_con}, with $N=100$. The object of interest to be covered is described by the Gaussian function $f(x,y) = A\exp\left(-\left(\frac{(x-x_o)^2}{2\sigma^2_x}+\frac{(y-y_o)^2}{2\sigma^2_y}\right)\right)$, with $(x_o,y_o) = (45, 45)$, $\sigma^2_x=\sigma^2_y = 80$, and $A = 40$, triangulated into  $|\mathcal{K}| = 338$ triangular facets from a point-cloud of $|\mathcal{P}|=200$ points. The planning horizon in the following experiments is set as $T=5$, and the mission time is $100$ time-steps. The term $\gamma(\tau)$ in the objective function of Eq. \eqref{eq:j1} is given by $\gamma(\tau) = \exp(T-\tau),~ \tau \in \{0,..,T-1\}$, $\delta=10$m, and $\omega=0.1$. Finally we should mention that we have used the Gurobi MIQP solver \cite{Gurobi} to solve the optimization problem (P2), on a Mac Studio M1 Ultra.

\begin{figure}
	\centering
	\includegraphics[width=\columnwidth]{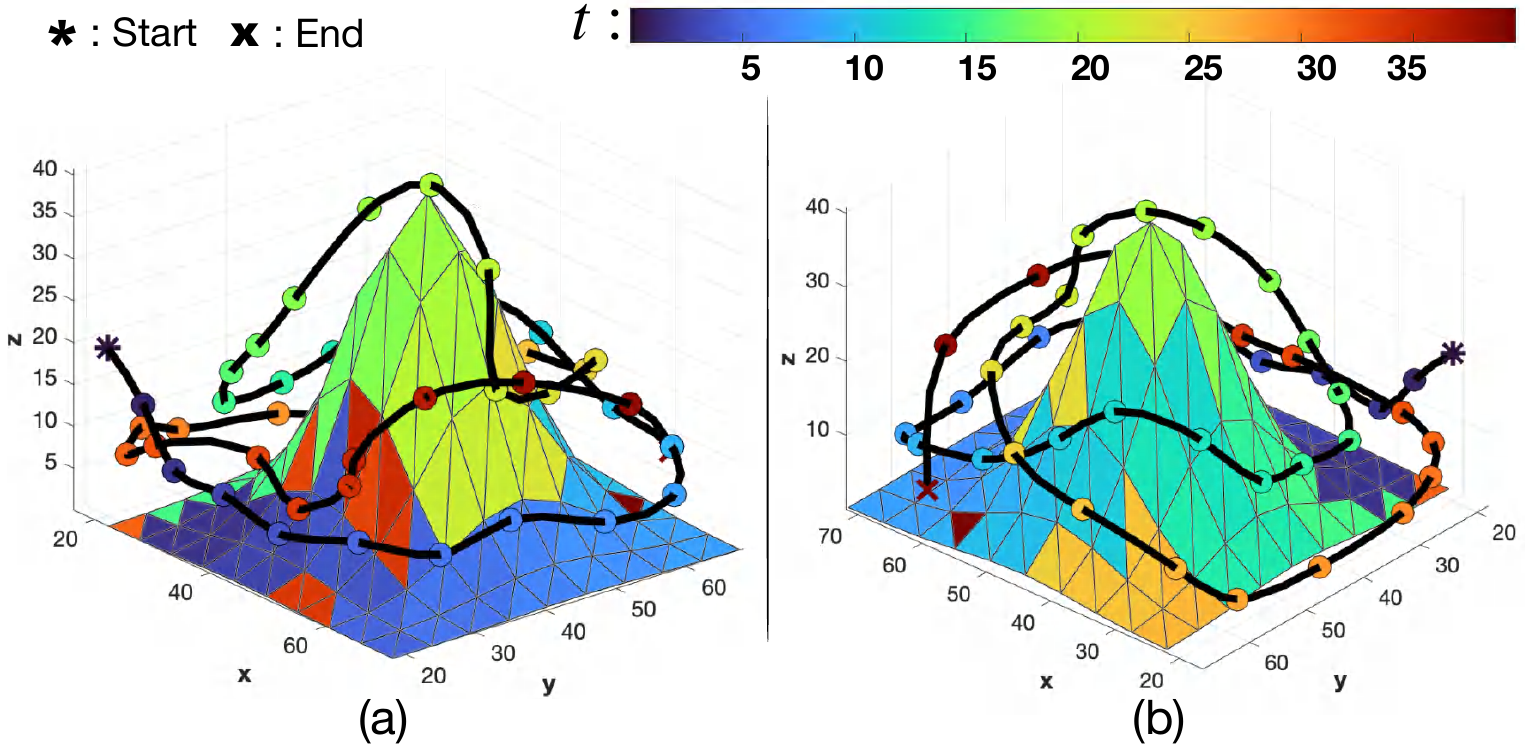}
	\caption{The figure illustrates the agent's trajectory which was used to cover the total surface area of the object of interest. (a) Front view of the object. (b) Back view of the object.}
	\label{fig:res3}
	\vspace{-3mm}
\end{figure}

We start the evaluation of the proposed approach with an illustrative example shown in Fig. \ref{fig:res1}. In order to simplify the analysis and for visual clarity, in this scenario we require that only a subset of facets $\tilde{\mathcal{K}}\subseteq \mathcal{K}$ need to be covered. For this reason, we have randomly sampled $|\tilde{\mathcal{K}}|=15$ facets from the object's surface as shown in Fig. \ref{fig:res1}(a) with purple color. Therefore, given the facets $\tilde{\mathcal{K}}$, in this scenario we seek to find the agent's optimal control inputs (i.e., kinematic and camera inputs) which minimize the coverage objective in Eq. \eqref{eq:objective_P2},  and respect the constraints shown in Eq. \eqref{eq:P2_1} -  Eq. \eqref{eq:P2_12}. Then, Problem (P2) is executed iteratively until all facets are covered. Figure \ref{fig:res1}(b) shows the resulting trajectory that the agent has executed in order to cover all facets. The agent starts at $(x,y,z) = (10,50,20)$ and finishes at $(x,y,z) = (48,23,22)$, with the start and final location shown as $\ast$ and $\times$ respectively. The facets to be covered and the agent states (shown as $\circ$) are color-coded according to the coverage elapsed time as shown in Fig. \ref{fig:res1}(b). Then, Fig. \ref{fig:res1}(c) shows the coverage trajectory in top-down view, and Fig. \ref{fig:res1}(d) shows the camera FOV configurations which have been used during the coverage mission. For visual clarity, we only show the FOV configurations which have resulted in a coverage event. When there is no coverage event the camera's FOV takes its initial state i.e., facing downwards. The zoom functionality is depicted with the camera FOVs colored in cyan, as shown in Fig. \ref{fig:res1}(d). In this scenario the coverage mission was executed in 17 time-steps, no duplication of work occurred, and all facets have been covered according to their visibility constraints.

\begin{figure}
	\centering
	\includegraphics[width=\columnwidth]{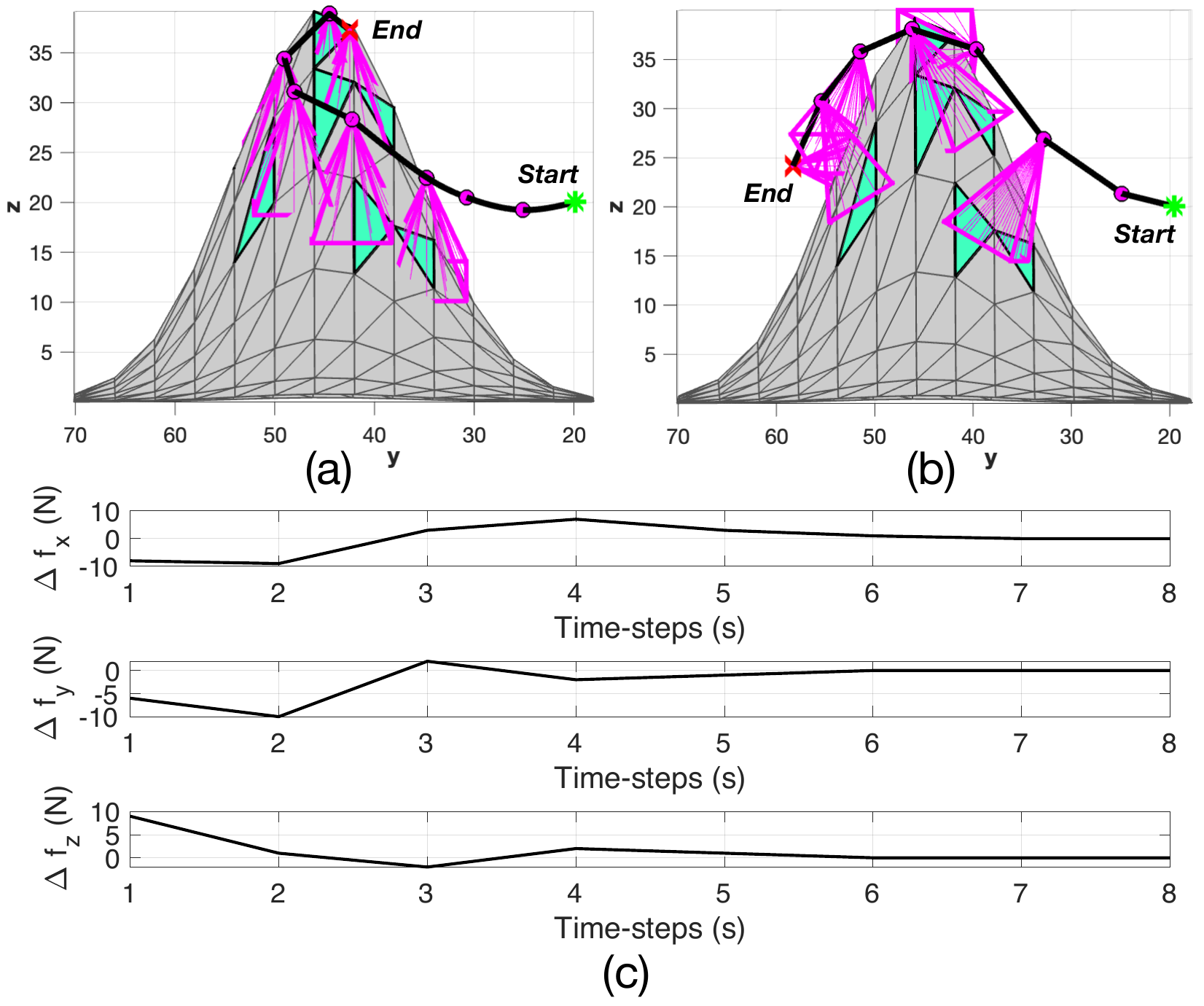}
	\caption{The effect of different objectives on the coverage trajectory. (a) minimizing the camera gimbal utilization, (b)(c) minimizing the deviations between consecutive motion control inputs.}
	\label{fig:res5}
	\vspace{-3mm}
\end{figure}

Figure \ref{fig:res2} shows the controller's output at time-step 4. More specifically, the figure shows the agent's planned coverage trajectory during the horizon $T$ of length 5 time-steps i.e., $x_{t+\tau+1|t}, t=4, \tau \in \{0,..,4\}$ marked with blue circles, and the predicted FOV configurations which are selected in order to maximize coverage which are also shown in blue color along with the covered facets. The agent's executed trajectory is $x_{1:4}$, and executed FOV states are shown in purple color along with the covered facets. Finally the agent's next state is shown in green color. The FOV states which result in no coverage are omitted from this figure for visual clarity. 

Next, Fig. \ref{fig:res3} shows the coverage trajectory which was executed for covering the total surface area of the object of interest. Fig. \ref{fig:res3}(a) shows the front view of the object, whereas  Fig. \ref{fig:res3}(b) shows the back view. The agent states (shown in $\circ$) and the facets are color-coded according to the coverage time as shown in the figure.  In this experiment, total coverage is achieved in 44 time-steps.

Figure \ref{fig:res5} shows that various secondary objectives can be incorporated into the proposed 3D coverage controller in order to achieve specific sub-goals. For instance we can design a sub-objective that aims in minimizing the rotations of the onboard camera (i.e., minimize the gimbal utilization), which for instance can increase the device's lifespan. Effectively, we can define this objective as $\mathcal{G}_\text{FOV} = \sum_{\tau=1}^{T-1}  \sum_{\hat{m}=1}^{|\mathcal{M}|} \left(s_{\hat{m},t+\tau+1|t} - s_{\hat{m},t+\tau|t} \right)$, where as discussed in Sec. \ref{ssec:controller}, the binary variable $s_{\hat{m},t+\tau+1|t}$ indicates the active FOV state (one out of $|\mathcal{M}|$) at time-step $t+\tau+1|t$. This particular sub-objective can now be combined with the main cost function $\mathcal{G}$ as a minimization of a weighted multi-objective cost i.e.,:
\begin{equation}\label{eq:sub_objective1}
	\min \left(\mathcal{G} + \hat{\omega} \mathcal{G}_\text{FOV} \right)
\end{equation}

\noindent where the parameter $\hat{\omega}$ controls the emphasis given to the secondary objective. Figure \ref{fig:res5}(a) shows effect of this sub-objective i.e., Eq. \eqref{eq:sub_objective1}, on the agent's coverage trajectory over a planning horizon of 8 time-steps, with $\hat{\omega}=10$. As illustrated in the figure, the agent managed to cover all green-colored facets without rotating its camera's FOV. Effectively, we can say that this coverage trajectory emulates the coverage trajectory obtained with a UAV equipped with a fixed and uncontrollable camera sensor. Observe now from Fig. \ref{fig:res5}(b) how the coverage trajectory for the same scenario changes when we minimize the control effort which is defined here as the sum of squared deviations between consecutive kinematic control inputs i.e., $\mathcal{G}_\text{motion} = \sum_{\tau=1}^{T-1}  ||u^f_{t+\tau|t} -  u^f_{t+\tau-1|t}||_2^2$. The coverage multi-objective cost function which now becomes $\min (\mathcal{G} + \hat{\omega} \mathcal{G}_\text{motion})$, is particular useful in designing energy efficient coverage trajectories which minimize sudden changes in the direction and speed of the agent, by driving $\mathcal{G}_\text{motion} \rightarrow 0$ as shown in Fig. \ref{fig:res5}(c). In Fig.  \ref{fig:res5}(b)(c) $\hat{\omega}$ has been set to 0.1, and the horizon length is equal to $T=8$ time-steps. In both scenarios shown in Fig. \ref{fig:res5} the agent starts from the same location, and is tasked to cover the same facets which are shown in green color. To summarize, we have shown how different secondary objectives can be incorporated into the main coverage cost function in order to capture various mission specifications. These secondary objectives can be weighted based on their priorities in order to meet the mission requirements.

\begin{figure}
	\centering
	\includegraphics[width=\columnwidth]{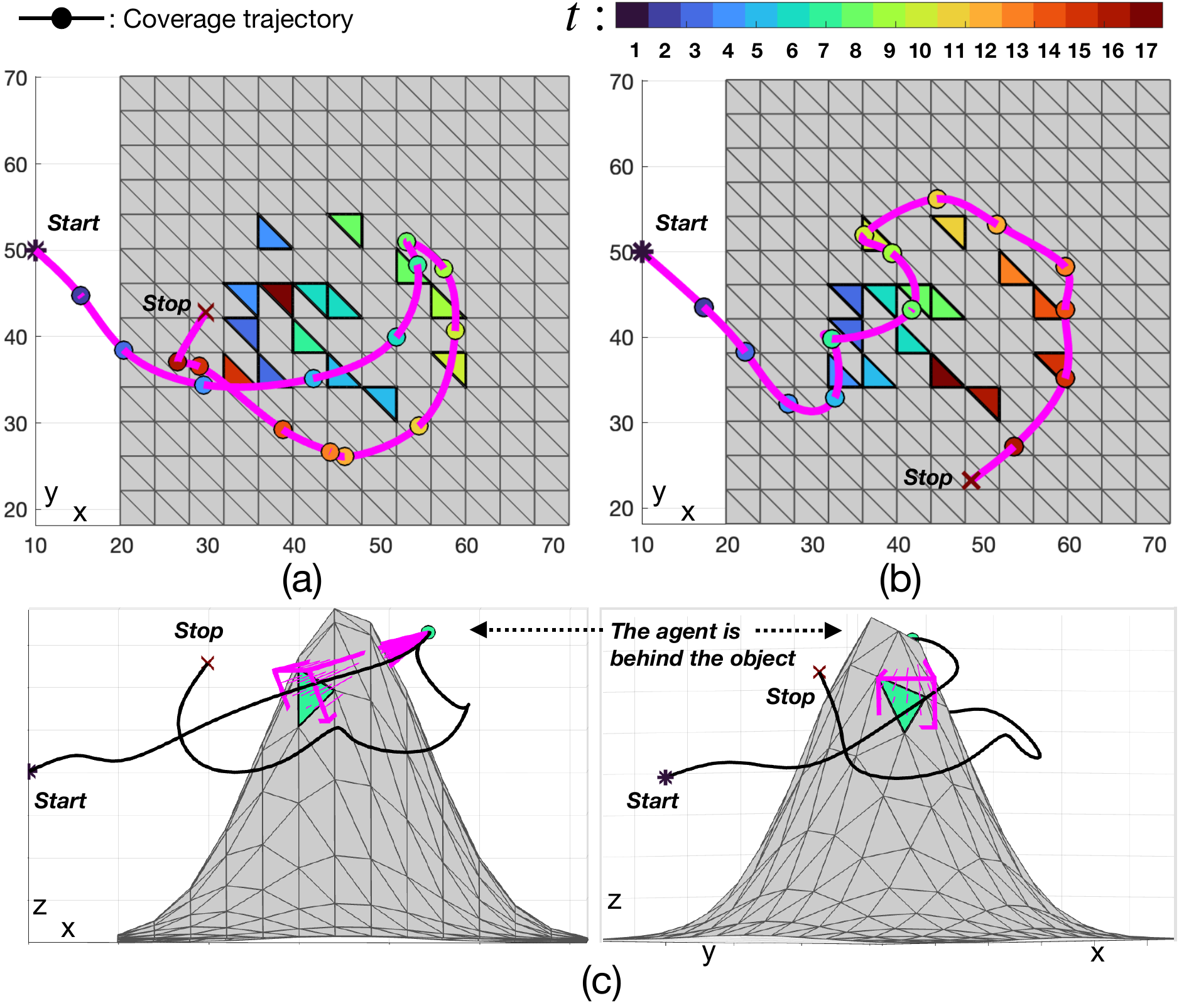}
	\caption{The effect of ray-tracing and visibility constraints on the coverage planning performance. (a) trajectory without ray-tracing, (b) trajectory with ray-tracing, (c) trajectory without ray-tracing: illustrative example of an erroneous coverage event which occurs at $t=7$, when visibility constraints are not enabled. The green facet is erroneously marked as covered, although it is not actually visible from the agent's location shown with a green circle.}
	\label{fig:res4}
	\vspace{-3mm}
\end{figure}

\begin{figure*}
	\centering
	\includegraphics[width=\textwidth]{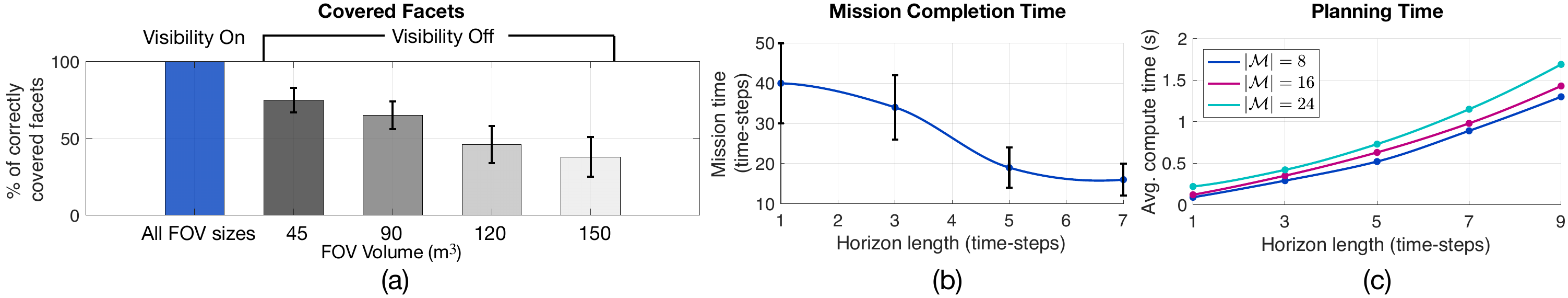}
	\caption{The figure shows: (a) The percentage of covered facets as a function of the FOV size, with visibility determination turned on and off, (b) the time required to complete a coverage mission (in time-steps) as a function of the horizon length, and (c) the computational time complexity analysis showing the average time required to compute a coverage plan as a function of the horizon length and the number of camera control inputs.}
	\label{fig:res7}
	\vspace{-2mm}
\end{figure*}

The majority of coverage planning approaches in the literature primarily focus on 2D environments \cite{Araujo2013, Cabreira2019s, Xie2020, TAES}. Approaches that address 3D environments often impose restrictive assumptions on the geometry of objects to be covered, such as cuboid-like structures \cite{ICUAS2022, PapaioannouTSMC, Tan2021}. In contrast, the proposed approach addresses the challenge of coverage planning in 3D environments for objects of arbitrary shapes, which can be represented as triangular meshes. In the proposed approach the agent operates in 3D space, and the object of interest for coverage is also fully defined in 3D.
Additionally, many existing works simplify the coverage planning problem to path planning \cite{Galceran2013, Collins2021, Jones2023} and also neglect the joint optimization of kinematic and camera control inputs \cite{Torres2016, kan2020online, Elmokadem2019}. In contrast, this work addresses the integrated problem of planning and control, and designs a predictive controller that simultaneously optimizes both kinematic and camera control inputs.
Furthermore, the proposed approach incorporates visibility determination constraints directly into the control loop, enabling long-horizon planning and optimal solutions. In contrast, most existing approaches address the problem using metaheuristic methods which do not guarantee finding an optimal solution, lack visibility determination, and are limited to myopic planning \cite{Tan2021, apostolidis2022cooperative, Jing2020, Sunan2017, ZachariaMED, Khan2017}.

To evaluate the performance of the proposed approach in comparison to existing methods, we focus on two key aspects: (a) emphasizing the significance of visibility determination by comparing coverage performance with and without incorporating visibility constraints, effectively illustrating how methods lacking visibility consideration are likely to perform, and (b) examining the influence of planning horizon length on coverage performance, thereby demonstrating the effectiveness of the proposed approach over methods that rely on myopic planning. Finally, (c), we discuss the computational complexity of the proposed approach.

\textbf{(a) Visibility determination:} To assess the importance of visibility determination, we evaluate the proposed approach both with and without the visibility constraints derived via ray tracing. Specifically, we remove the binary variables $b_{\hat{\varpi},\hat{\kappa}}$ and $b^{\tilde{\mathcal{A}}}_{\hat{\varpi},t+\tau+1|t}$ from Eq. \eqref{eq:P2_9}, resulting in the following modified coverage constraint: $\bar{b}_{\hat{\kappa},\hat{m},t+\tau+1|t} = s_{\hat{m},t+\tau+1|t}~\wedge~b^{\mathcal{V}}_{\hat{\kappa},\hat{m},t+\tau+1|t}$. The results demonstrate that, although this modified constraint directs the agent toward covering the facets of the object of interest, it does not account for whether a facet $\kappa$, located within the agent's camera FOV, is actually visible (i.e., whether a light ray can trace back to $\kappa$). Figure \ref{fig:res4} illustrates the coverage trajectories of the object of interest with and without visibility constraints, highlighting the impact of incorporating visibility determination. Specifically, Fig. \ref{fig:res4}(a) provides a top-down view of the agent's coverage trajectory without visibility constraints (i.e., no ray-tracing), while Fig. \ref{fig:res4}(b) depicts the coverage trajectory obtained with visibility constraints enabled. In Fig. \ref{fig:res4}(a), the agent is unable to determine whether a particular facet $\kappa$ within its camera field of view (FOV) is actually visible. As a result, many facets that are not visible but merely fall within the camera FOV are erroneously marked as covered, leading to an inaccurate coverage trajectory, as shown in the figure. In this specific scenario, the agent's camera FOV spans across the object's body. Without visibility constraints, the agent cannot differentiate which facets belong to the foreground (i.e., the visible surface area of the object), as illustrated in Fig. \ref{fig:res4}(c).

To further assess the importance of visibility determination in identifying visible facets for coverage, a Monte Carlo simulation comprising 100 trials was performed. In each trial, the agent was randomly initialized within the simulated environment depicted in Fig. \ref{fig:res1}, and the proposed controller was executed both with and without visibility constraints. Coverage facets were randomly selected from the set $\mathcal{K}$, with the number of facets sampled uniformly from the range [10, 20]. The results, shown in Fig. \ref{fig:res7}(a), for various field-of-view (FOV) sizes (measured in volume), demonstrate that when visibility determination is enabled, the controller achieves full coverage of all facets, irrespective of FOV size, without any degradation in performance. This outcome is attributed to the integration of ray-tracing in the proposed method, which ensures accurate identification of visible facets. Conversely, disabling visibility determination leads to a noticeable performance decline as the FOV size increases. This reduction occurs because larger FOVs encompass facets within the convex hull of the FOV that are not genuinely visible. In contrast, smaller FOVs necessitate closer proximity to the object of interest, inherently aligning the agent's view with the visible facets. Larger FOVs, however, may include a broader set of facets, some of which remain occluded despite being within the FOV. These findings highlight the critical role of visibility determination as a fundamental component in effective coverage planning, a feature that is currently absent in many existing approaches.

\textbf{(b) Planning horizon:}  Figure \ref{fig:res7}(b) illustrates the relationship between the planning horizon length and the average mission completion time. These results were obtained through a Monte Carlo simulation consisting of 100 trials, where the agent was randomly initialized within the environment depicted in Fig. \ref{fig:res1}. The objective in each trial was to cover 15 facets randomly selected from the object's surface mesh. The figure reveals that for a planning horizon of  $T=1$  (i.e., myopic planning), the average mission completion time is approximately 40 time-steps. However, as the planning horizon length increases, the mission completion time decreases significantly, stabilizing at around 16 time-steps for  $T=6$  or greater, i.e., $60\%$ improvement in mission completion time. While this behavior is inherently dependent on the specific scenario and problem, it clearly demonstrates the importance of long-horizon planning for enhancing coverage performance. Longer planning horizons enable the agent to anticipate future control inputs, thereby optimizing the mission objective, as defined by Eq. \eqref{eq:j1}. Conversely, myopic approaches, which lack the ability to predict and incorporate future states into the decision-making process, result in reduced performance and longer mission durations.\\

\textbf{(c) Computational complexity:} The coverage planning problem is formulated as a rolling finite-horizon optimal control problem (FHOCP) and solved using mixed-integer quadratic programming (MIQP). It is important to emphasize that the proposed approach is optimal in the sense that it computes the control inputs that minimize the objective function (i.e., Eq. \eqref{eq:j1}). This is because Problem (P2) is an MIQP with a convex objective function and linear constraints, ensuring the existence of a unique global minimum for any continuous relaxation of the problem. MIQP solvers, utilizing branch-and-bound or branch-and-cut algorithms, systematically explore subsets of the feasible space by enumerating feasible integer solutions. These solvers leverage convexity to prune suboptimal branches, guaranteeing convergence to the global minimum-unlike existing approaches that rely on heuristics.
While modern solvers are highly effective at solving MIQPs, combining theoretical guarantees with heuristic accelerations and exploiting problem structure and convexity to efficiently find the optimal solution, the computational complexity of MIQP problems can still grow exponentially in the worst case with respect to the number of decision variables. To evaluate this, we measured the average time required for an agent to compute a coverage plan during the mission depicted in Fig. \ref{fig:res1} (where the agent is randomly initialized in the environment and tasked with covering  $n$  facets randomly sampled from the range [10, 20]). 

Figure \ref{fig:res7}(c) illustrates the average time required to execute one iteration of Problem (P2) (i.e., generating a coverage trajectory) over 100 random trials, as a function of the planning horizon length ($T$) and the number of camera control inputs ($|\mathcal{M}|$). As shown, the computational complexity increases as the planning horizon length and the number of camera control inputs grow. The results confirm the theoretical properties of the problem while demonstrating that certain problem sizes can be solved efficiently to optimality. Additionally, the results presented in Fig. \ref{fig:res7}(b) and Fig. \ref{fig:res7}(c) offer valuable insights into the trade-offs between performance and computational complexity. These findings can be leveraged to fine-tune the controller, enabling it to meet specific mission requirements effectively. It is worth noting that recent advances in optimization methods, such as adaptive neighborhood search techniques \cite{Hendel2021} and machine learning-based approaches \cite{bertsimas2022online}, enable large MIQP problems to be solved efficiently and in real-time. We should note here that in real-world settings, it is crucial to ensure that the time required for computing a coverage plan is either guaranteed, or can be accurately estimated. For this reason, future work will focus on this aspect, as well as on developing strategies to handle computation failures or timeouts encountered during optimization.

\begin{figure}
	\centering
	\includegraphics[width=\columnwidth]{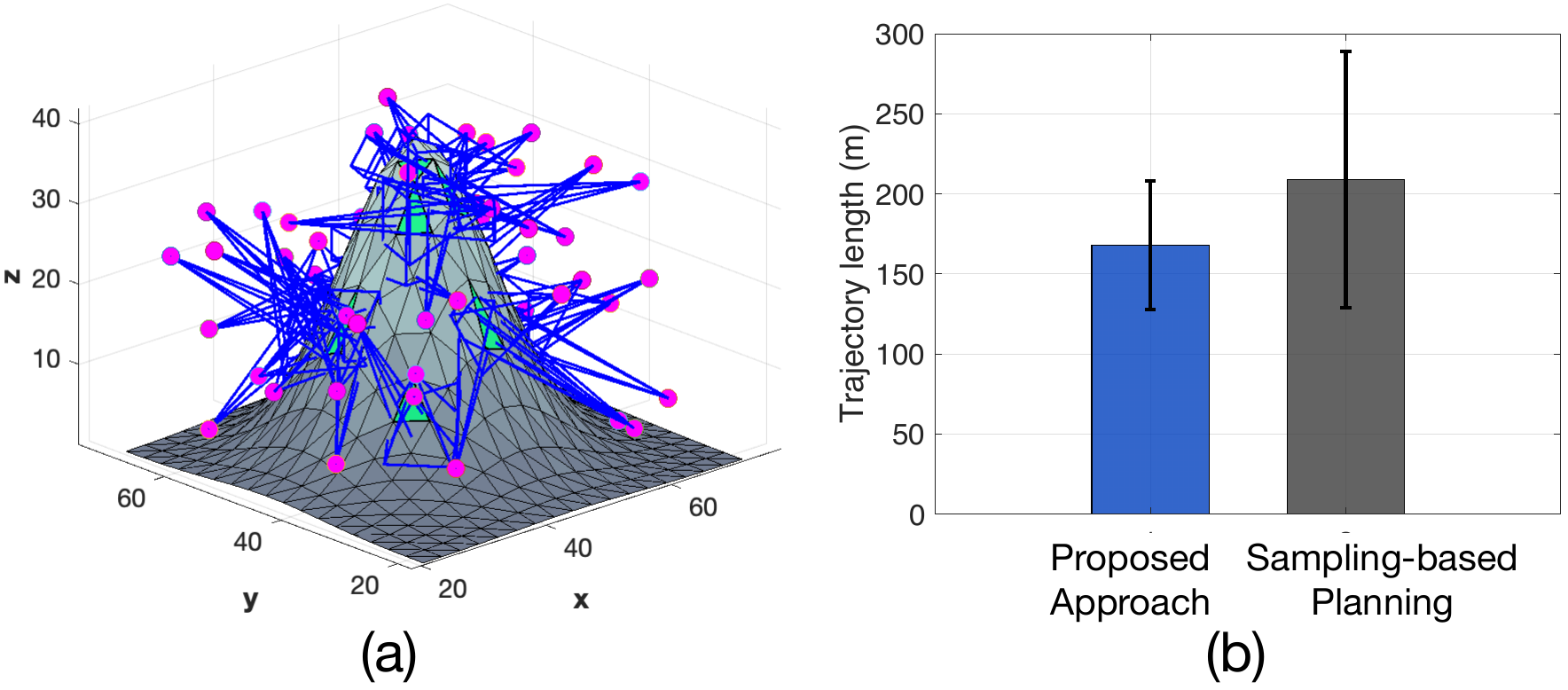}
	\caption{(a) Illustrative example of sampled viewpoint configurations, (b) Performance comparison of the proposed approach with sampling-based view planning methods.}
	\label{fig:res_x}
	\vspace{-5mm}
\end{figure}

\color{black}
Finally, we compare the proposed approach with the sampling-based view-planning methodology i.e., a popular variant of the coverage path planning problem \cite{maboudi2023review}. Sampling-based view-planning methods usually utilize a two-stage approach. First, candidate viewpoints are sampled within a constrained space around the target object (e.g., Fig.~\ref{fig:res_x}(a)). In the second stage, a subset of these viewpoints is selected by solving an approximate version of the set covering problem (SCP) using heuristics such as greedy search, genetic algorithms, or hybrid methods. The view planning problem is nominally formulated as the following constrained optimization problem: 
\begin{equation}\label{eq:Sampling}
\begin{aligned}
	\min \quad & \sum_{j=1}^{N} \psi_j \\
	\text{s.t.} \quad & \Omega \cdot \bm{\psi} \geq 1, \\
	& d(j,l) \leq \Delta_t \|\mathbf{v}_{\text{max}}\|, \quad \forall j \neq l, \\
	& \psi_j \in \{0,1\}, \quad \forall j, \\
	&\Omega(i,j) \in \{0,1\}, \quad \forall i,j, \\
	& i \in \{1, \dots, M\}, ~ j,l \in \{1, \dots, N\}.
\end{aligned}
\end{equation}

In Eq.\eqref{eq:Sampling}, $\psi_j$ is a binary decision variable indicating whether the $j_\text{th}$ viewpoint is selected. The binary visibility matrix $\Omega \in \{0,1\}^{M \times N}$ defines visibility constraints, where $M$ is the number of object surface facets requiring coverage, and $N$ is the number of sampled viewpoints. If $\Omega(i,j) = 1$, facet $i$ is visible and covered by viewpoint $j$. The constraint on the distance $d(j,l)$ ensures a minimum distance between connected viewpoints $i$ and $j$, determined by the UAV's allowed maximum velocity $\mathbf{v}_{\text{max}}$.

To evaluate our approach against the sampling-based view-planning technique i.e., Eq. \eqref{eq:Sampling}, we conducted 40 Monte Carlo simulations, randomly selecting 10 facets from the object's surface (using the simulation setup described and shown in Fig. \ref{fig:res1}) and varying the UAV's initial position, measuring the length of the coverage trajectory as an indication of the coverage performance. Unlike the proposed approach, which directly optimizes the UAV trajectory, view-planning methods generate a coverage path (i.e., a sequence of connected viewpoints). In order to compare the two approaches, we convert the coverage path $P$, generated by the view-planning approach, into a feasible UAV trajectory by solving the following receding horizon control problem over a rolling finite horizon of length $H$ time-steps:
\begin{equation}\label{eq:MPC}
\begin{aligned}
    \min \quad & \sum_{t=0}^{H-1}  \| x_t^{\mathbf{p}} - \hat{P}_t \|^2  \\
    \text{s.t.} \quad & x_{t+1} = A x_t + B f_t, \quad \forall t \in \{0, \dots, H-1\}, \\
    & x_0 = x_{\text{init}}, \\
    & x_t \in \mathcal{X}, ~f_t \in \mathcal{F}, \quad \forall t \in \{0, \dots, H-1\}.
\end{aligned}
\end{equation}

\noindent In essence we want to find the UAV control inputs $\{f_0,\ldots,f_{H-1}\}$ in a rolling horizon fashion (we have used $H=5$) that optimally track the reference path $\hat{P}_t$. Here, $\hat{P}_t$ is a continuous reference path obtained from $P$ via spline interpolation \cite{wang2002arc}. 
Figure~\ref{fig:res_x}(a) illustrates an example of the sampled viewpoint configurations (i.e., UAV positions and camera orientations) generated in the first stage of the view-planning approach. In each simulation, we sample 200 valid viewpoints around the object of interest. The coverage path is then computed by solving Eq.~\eqref{eq:Sampling} using a greedy local search strategy \cite{jing2016view}. Figure \ref{fig:res_x}(b) illustrates the average trajectory length obtained using the two approaches, demonstrating that the UAV coverage trajectory generated by the proposed approach is approximately 25\% more efficient (i.e., shorter in terms of distance) compared to that of the view planning approach. The proposed method jointly optimizes the UAV's motion and camera control inputs under visibility constraints by solving an optimal control problem, whereas the view planning approach approximates this behavior through sampling.

\color{black}

\begin{figure*}
	\centering
	\includegraphics[width=\textwidth]{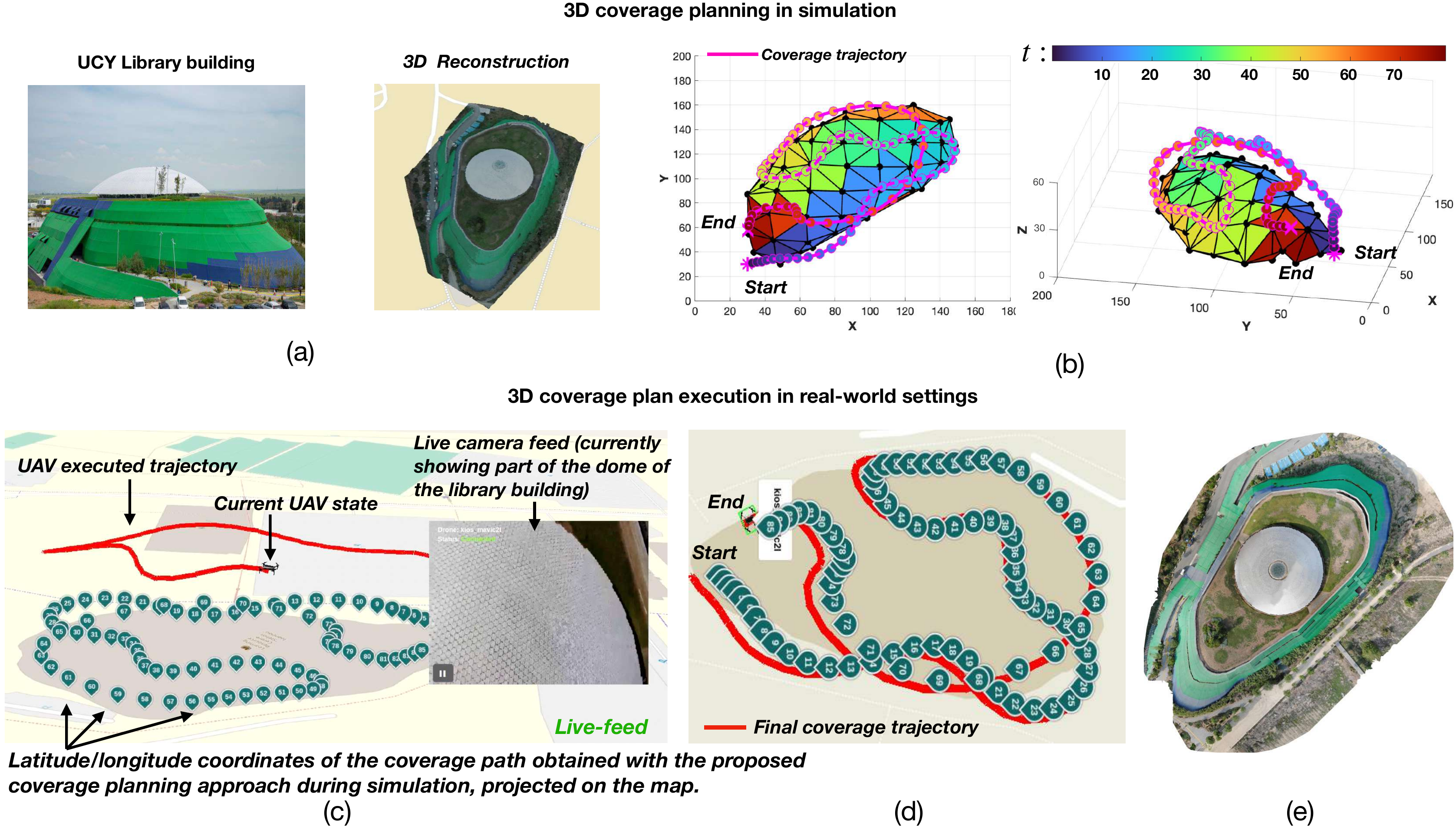}
	\caption{Real-world 3D coverage mission. (a) Object of interest, (b) Coverage trajectory obtained with the proposed approach, (c)-(e) The generated coverage plan executed in real-world settings.}
	\label{fig:res6b}
	\vspace{-5mm}
\end{figure*}

\color{black}

\subsection{Real-world Experiment}
In the second part of the evaluation, we demonstrate the performance of the proposed approach in a real-world coverage mission. More specifically, the main objective here is to compute the 3D coverage plan for the Library building of the University of Cyprus. The library building, shown in Fig. \ref{fig:res6b}(a), is an oval-like structure with major and minor axis diameters of about 170m and 85m respectively, and a height of about 30m. As discussed in Sec. \ref{ssec:object_of_interest}, as a first step we have used 3D reconstruction techniques to build a 3D model of the library structure i.e., the library building was first reconstructed as a point-cloud using Open3D \cite{Zhou2018}, as shown in Fig. \ref{fig:res6b}(a), and then converted into a triangle mesh using Delaunay triangulation. As a result, a total of 82 triangles were use to model the surface of the library building as shown in Fig. \ref{fig:res6b}(b). 

For our experimental evaluation we have used the DJI Mavic Enterprise consumer drone, equipped with a 12MP gimballed camera system, with horizontal and vertical rotation angles in the ranges $\Phi=\{300, 315, 0, 45, 60\}$deg, and $\Theta=\{0, 30, 90\}$deg respectively, and a square FOV footprint, at a hight of approximately 18m, equal to 20m-by-20m. The zoom functionality was not utilized for this test i.e., $Z=\{1\}$, and the rest of the parameters have been set as discussed in Sec. \ref{ssec:simtest}. Figure \ref{fig:res6b}(b) shows the generated coverage trajectory obtained with the proposed technique. In this figure the UAV's trajectory is color-coded based on the mission elapsed time. Similarly, the object's facets are colored based on the time-step which have been observed by the UAV (i.e., based on the coverage-time). 

 As illustrated in the figure the generated trajectory enabled the coverage of the total surface area of the structure (the camera FOV configurations are omitted for visual clarity). Subsequently in order to verify the generated coverage plan in practice, we have utilized our automated drone tasking platform \cite{Terzi2019} to command the UAV to execute the coverage trajectory generated with the proposed approach. Figure \ref{fig:res6b}(c) shows an instance of the executed coverage plan during flight-time. Specifically, the figure shows a snapshot of the live-feed obtained from our mission control platform while executing the coverage trajectory. Figure \ref{fig:res6b}(d) shows in top-down view of the executed coverage trajectory in red. The green waypoints shown in the figure correspond to the latitude-longitude coordinates of the planned trajectory computed with the proposed approach during simulation. Finally, Fig. \ref{fig:res6b}(e) shows the building's orthophotomap generated from the images acquired during the real-world mission, indicating the covered surface area of the structure i.e., the proposed approach achieves full coverage by minimizing the cost function shown in Eq. \eqref{eq:j1}.

\textbf{Implementation details and lessons learned:} The UAV was operated using the DJI Mobile Software Development Kit (SDK), which facilitated the creation of a custom Android mobile application installed on a smartphone paired with the UAV. This smartphone acted as an intermediary by wirelessly connecting to the UAV's remote controller, enabling the transmission of commands to the UAV and the reception of telemetry data such as GPS coordinates, altitude, battery status, speed, and sensor readings. The proposed coverage controller was implemented and executed on a ground control station (GCS) using Matlab. The GCS computed the UAV trajectories and transmitted them to the drone via the mobile application on the Android smartphone connected to the remote controller. Telemetry data from the UAV was also relayed back to the GCS through this setup. Communication between the GCS and the UAV was facilitated by a VPN server over a wireless network, allowing data exchange between the two. Mission monitoring was conducted live using a custom-built UAV tasking platform \cite{Terzi2019}. 

While this prototype setup demonstrated the effectiveness of the proposed approach, it highlighted several areas for future improvement. GPS-based positioning introduced localization errors and inconsistencies, leading to distorted coverage plans in certain scenarios. Additionally, network latency resulted in communication delays that impacted the drone's operation, and environmental disturbances caused deviations from the planned trajectories. Future work will aim to address these challenges by enhancing the approach to incorporate stochastic and robust predictive control methods for managing environmental uncertainties, integrating the controller with the UAV's onboard systems to improve real-time responsiveness, and scaling the methodology to support multiple UAV agents.

\color{black}

\section{Conclusion} \label{sec:conclusion}

In this work we propose a jointly-optimized trajectory generation and camera control approach for automated coverage planning in 3D environments. The proposed coverage planning approach integrates ray-tracing into the coverage planning process thus allowing an autonomous mobile agent to determine the visible parts of the object of interest; and generate look-ahead coverage trajectories by jointly optimizing its kinematic and camera control inputs over a rolling finite planning horizon. We show how the coverage planning problem can be formulated as a finite horizon optimal control problem, and then solved using mixed integer programming. Finally, the performance of the proposed approach is demonstrated through extensive synthetic and real-world experiments. 

\section*{Acknowledgments}
The project is implemented under the Border Management and Visa Policy Instrument (BMVI) and is co-financed by the European Union and the Republic of Cyprus (GA:BMVI/2021-2022/SA/1.2.1/015), and supported by the GLIMPSE project EXCELLENCE/0421/0586 co-financed by the European Regional Development Fund and the Republic of Cyprus through the Research and Innovation Foundation's RESTART 2016-2020 Programme for Research, Technological Development and Innovation, and by the European Union's Horizon 2020 research and innovation programme under grant agreement No 739551 (KIOS CoE).

\bibliographystyle{IEEEtran}
\bibliography{IEEEabrv,main} 
\begin{IEEEbiography}[{\includegraphics[width=1in,height=1.25in,clip,keepaspectratio]{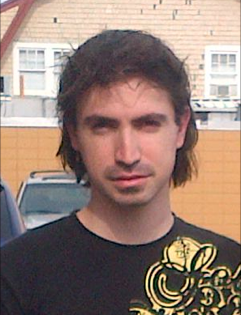}}]%
{Savvas Papaioannou} is a Senior Research Scientist at the KIOS Research and Innovation Center of Excellence (KIOS CoE) at the University of Cyprus (UCY). He earned his B.Sc. degree in Electronic and Computer Engineering from the Technical University of Crete (Greece) in 2011, his M.Sc. degree in Electrical Engineering from Yale University (USA) in 2013, and his DPhil (Ph.D.) in Computer Science from the University of Oxford (UK) in 2017.
Dr. Papaioannou's research interests focus on the modeling, estimation, control, and optimization of autonomous intelligent systems. His work spans multi-agent systems, intelligent unmanned aerial vehicles (UAVs), multi-target tracking, artificial intelligence, and their applications.
His work has received several awards, including the Best Paper Award at the 10th IEEE Symposium on Embedded Systems for Real-time Multimedia (ESTIMedia) in 2012, the First Prize in the Cooperative Aerial Robots Inspection Challenge at the 62nd IEEE Conference on Decision and Control (CDC) in 2023, and the Runner-up Award (2nd Prize) in the Multi-Robot Perception and Navigation Challenges in Logistics and Inspection Tasks competition at the 2024 IEEE/RSJ International Conference on Intelligent Robots and Systems (IROS).
Dr. Papaioannou serves as a reviewer for numerous IEEE and ACM conferences and journals. Additionally, he has been actively involved in the organization of international conferences, including the 2018 European Control Conference (ECC), the 2022 IFAC Symposium on Fault Detection, Supervision, and Safety for Technical Processes (Safeprocess), the 2023 Mediterranean Conference on Control and Automation (MED), and the 6th International Conference on Control and Fault-Tolerant Systems (SysTol) in 2025.
\end{IEEEbiography}
\vspace{-15mm}
\begin{IEEEbiography}[{\includegraphics[width=1in,height=1.25in,clip,keepaspectratio]{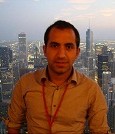}}]%
{Panayiotis Kolios} is currently a Research Assistant Professor at the KIOS Research and Innovation Centre of Excellence of the University of Cyprus. He received his BEng and PhD degree in Telecommunications Engineering from King's College London in 2008 and 2011, respectively. Before joining the KIOS CoE, he worked at the Department of Communications and Internet Studies at the Cyprus University of Technology and the Department of Computer Science of the University of Cyprus (UCY). His work focuses on both basic and applied research on networked intelligent systems. Some examples of systems that fall into the latter category include intelligent transportation systems, autonomous unmanned aerial systems and the plethora of cyber-physical systems that arise within the Internet of Things. Particular emphasis is given to emergency management in which natural disasters, technological faults and man-made attacks could cause disruptions that need to be effectively handled. Tools used include graph theoretic approaches, algorithmic development, mathematical and dynamic programming, as well as combinatorial optimization.
\end{IEEEbiography}
\vspace{-20mm}
\begin{IEEEbiography}[{\includegraphics[width=1in,height=1.25in,clip,keepaspectratio]{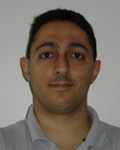}}]%
{Theocharis Theocharides} is an Associate Professor in the Department of Electrical and Computer Engineering, at the University of Cyprus and a faculty member of the KIOS Research and Innovation Center of Excellence where he serves as the Research Director. Theocharis received his Ph.D. in Computer Engineering from Penn State University, working in the areas of low-power computer architectures and reliable system design, where he was honored with the Robert M. Owens Memorial Scholarship, in May 2005. He has been with the Electrical and Computer Engineering department at the University of Cyprus since 2006, where he directs the Embedded and Application-Specific Systems-on-Chip Laboratory. His research focuses on the design, development, implementation, and deployment of low-power and reliable on-chip application-specific architectures, low-power VLSI design, real-time embedded systems design and exploration of energy-reliability trade-offs for Systems on Chip and Embedded Systems. His focus lies on acceleration of computer vision and artificial intelligence algorithms in hardware, geared towards edge computing, and in utilizing reconfigurable hardware towards self-aware, evolvable edge computing systems. His research has been funded by several National and European agencies and the industry, and he is currently involved in over ten funded ongoing research projects. He serves on several organizing and technical program committees of various conferences (currently serving as the Application Track Chair for the DATE Conference), is a Senior Member of the IEEE and a member of the ACM. He  is currently an Associate Editor for the ACM Transactions on Emerging Technologies in Computer Systems, IEEE Consumer Electronics magazine, IET's Computers and Digital Techniques, the ETRI journal and Springer Nature Computer Science. He also serves on the Editorial Board of IEEE Design \& Test magazine.
\end{IEEEbiography}
\vspace{-20mm}
\begin{IEEEbiography}[{\includegraphics[width=1in,height=1.25in,clip,keepaspectratio]{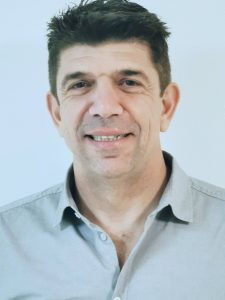}}]%
{Christos Panayiotou} is a Professor with the Electrical and Computer Engineering (ECE) Department at the University of Cyprus (UCY). He is also the Acting Director of the KIOS Research and Innovation Center of Excellence for which he is also a founding member. Christos has received a B.Sc. and a Ph.D. degree in Electrical and Computer Engineering from the University of Massachusetts at Amherst, in 1994 and 1999 respectively. He also received an MBA from the Isenberg School of Management, at the aforementioned university in 1999. Before joining the University of Cyprus in 2002, he was a Research Associate at the Center for Information and System Engineering (CISE) and the Manufacturing Engineering Department at Boston University (1999 - 2002). His research interests include modeling, control, optimization and performance evaluation of intelligent transportation systems, cyber-physical systems, discrete event and hybrid systems,  event detection and localization, fault diagnosis, machine learning, wireless, ad hoc and sensor networks, resource allocation, and intelligent buildings. Christos has published more than 340 papers in international refereed journals and conferences and is the recipient of the 2014 Best Paper Award for the journal Building and Environment (Elsevier). He is an Associate Editor for the IEEE Transactions of Intelligent Transportation Systems, and the Journal of Discrete Event Dynamical Systems. He had a long tenure on the Conference Editorial Board of the IEEE Control Systems Society (2001-2023) and served as an Associate Editor of the IEEE Transactions on Control Systems Technology during 2016-2020 and of the European Journal of Control during 2014-2021.  He held several positions in organizing committees and technical program committees of numerous international conferences, including General Chair of the 31st Mediterranean Conference on Control and Automation (MED2023), General Chair of the 23rd European Working Group on Transportation (EWGT2020), and General Co-Chair of the 2018 European Control Conference (ECC2018).
\end{IEEEbiography}
\vspace{-10mm}
\begin{IEEEbiography}[{\includegraphics[width=1in,height=1.25in,clip,keepaspectratio]{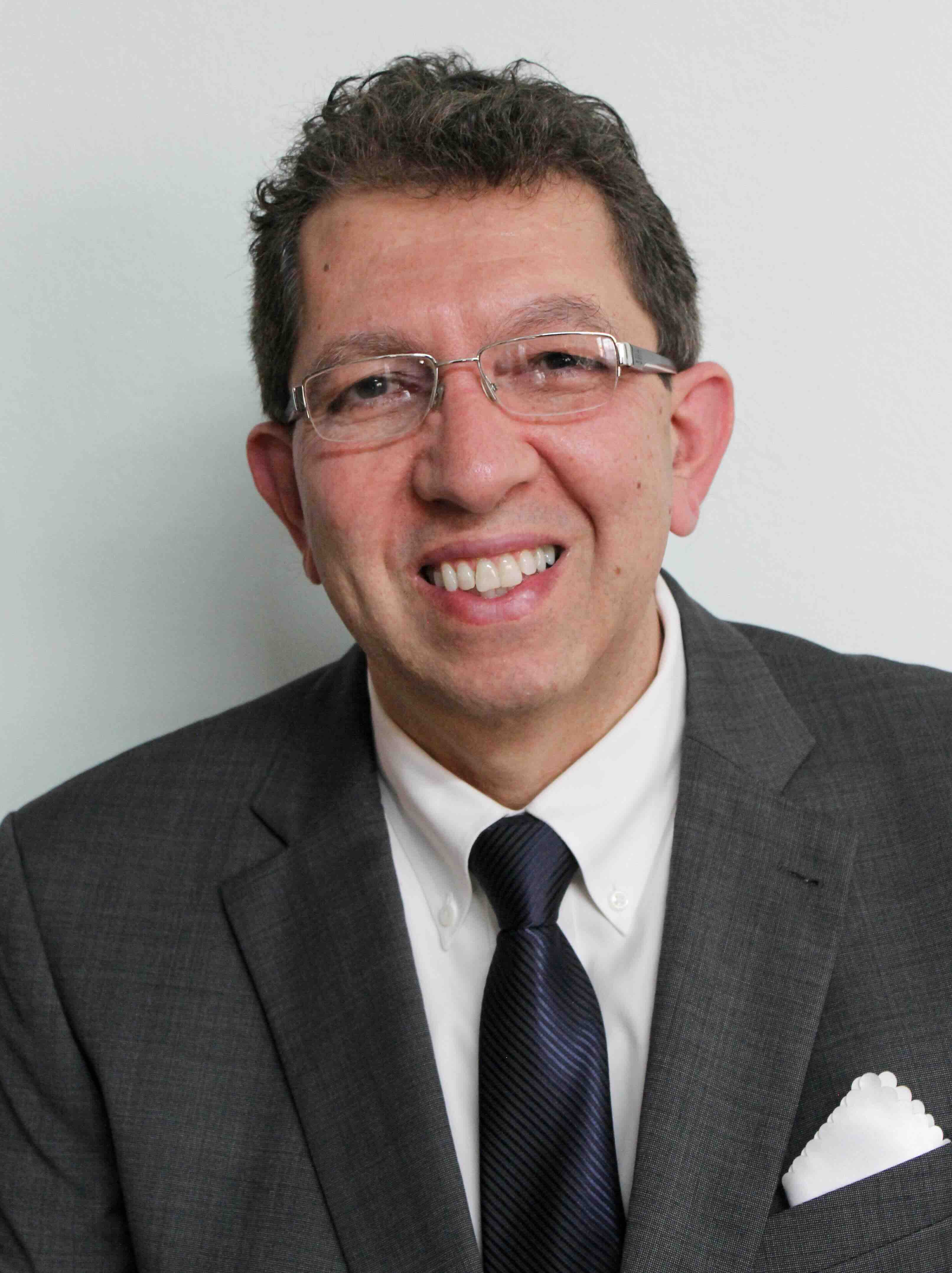}}]%
{Marios M. Polycarpou} is a Professor of Electrical and Computer Engineering and the Chair of the Advisory Board of the KIOS Research and Innovation Center of Excellence at the University of Cyprus. He is also a Founding Member of the Cyprus Academy of Sciences, Letters, and Arts, an Honorary Professor of Imperial College London, and a Member of Academia Europaea (The Academy of Europe).  He received the B.A degree in Computer Science and the B.Sc. in Electrical Engineering, both from Rice University, USA in 1987, and the M.S. and Ph.D. degrees in Electrical Engineering from the University of Southern California, in 1989 and 1992 respectively. His teaching and research interests are in intelligent systems and networks, adaptive and learning control systems, fault diagnosis, machine learning, and critical infrastructure systems.
Prof. Polycarpou is the recipient of the 2023 IEEE Frank Rosenblatt Technical Field Award and the 2016 IEEE Neural Networks Pioneer Award. He is a Fellow of IEEE and IFAC. He served as the President of the IEEE Computational Intelligence Society (2012-2013), as the President of the European Control Association (2017-2019), and as the Editor-in-Chief of the IEEE Transactions on Neural Networks and Learning Systems (2004-2010). Prof. Polycarpou currently serves on the Editorial Boards of the Proceedings of the IEEE and the Annual Reviews in Control. His research work has been funded by several agencies and industry in Europe and the United States, including the prestigious European Research Council (ERC) Advanced Grant, the ERC Synergy Grant and the EU-Widening Teaming program.
\end{IEEEbiography}

\flushbottom
\balance

\end{document}